\documentclass[runningheads]{llncs}
\usepackage[T1]{fontenc}
\usepackage{graphicx}
\usepackage[pagebackref,breaklinks,colorlinks]{hyperref}
\usepackage[many]{tcolorbox}
\usepackage[nameinlink]{cleveref}
\usepackage{multirow}
\usepackage[normalem]{ulem}
\usepackage{amssymb}

\usepackage{subcaption}
\usepackage[normalem]{ulem}
\usepackage[title]{appendix}
\useunder{\uline}{\ul}{}

\usepackage[font=small,skip=2pt]{caption}

\begin{document}
\title{Embedding Arithmetic: A Lightweight, Tuning-Free Framework for Post-hoc Bias Mitigation in Text-to-Image Models}

\author{Venkatesh Thirugnana Sambandham\inst{1}\orcidID{0000-0002-6974-7669} \and
Torsten Schön\inst{1}\orcidID{0000-0001-5763-3392}}
\authorrunning{V. T. Sambandham et al.}
%
\institute{AIMotion Bavaria, Technische Hochschule Ingolstadt, Esplanade 10, 85049 Ingolstadt, Germany
\email{\{venkatesh.thirugnanasambandham, torsten.schoen\}@thi.de}}
\maketitle              
\begin{abstract}
Modern text-to-image (T2I) models amplify harmful societal biases, challenging their ethical deployment.
We introduce an inference-time method that reliably mitigates social bias while keeping prompt semantics and visual context (background, layout, and style) intact. This ensures context persistency and provides a controllable parameter to adjust mitigation strength, giving practitioners fine-grained control over fairness-coherence trade-offs.
Using Embedding Arithmetic, we analyze how bias is structured in the embedding space and correct it without altering model weights, prompts, or datasets.
Experiments on FLUX 1.0-Dev and Stable Diffusion 3.5-Large show that the conditional embedding space forms a complex, entangled manifold rather than a grid of disentangled concepts.
To rigorously assess semantic preservation beyond the circularity and bias limitations of of CLIP scores, we propose the Concept Coherence Score (CCS).
Evaluated against this robust metric, our lightweight, tuning-free method significantly outperforms existing baselines in improving diversity while maintaining high concept coherence, effectively resolving the critical fairness-coherence trade-off. 
By characterizing how models represent social concepts, we establish geometric understanding of latent space as a principled path toward more transparent, controllable, and fair image generation. 
The complete source code will be released upon acceptance, a demo notebook with basic implementations can be found at \url{https://github.com/cvims/EMBEDDING-ARITHMETIC}
\end{abstract}

\section{Introduction}

Recent advancements in generative AI have led to text-to-image (T2I) models capable of producing high-quality, diverse, and creative visual content with remarkable quality~\cite{zhang2017stackgan,mirza2014conditional,rombach2022high,esser2024scaling,oppenlaender2022creativity,baldridge2024imagen}. 
While these models unlock new frontiers in art, design, and entertainment, their reliance on vast, web-scale datasets means they often inherit and amplify harmful societal biases~\cite{friedrich2025auditing,seshadri-etal-2024-bias}. 
When prompted with neutral terms like "a photo of a CEO," these systems frequently generate images that conform to narrow demographic stereotypes, posing significant risks to fairness and equitable representation~\cite{wan2025maleceofemaleassistant,luccioni2023stable}.

To address this critical issue, researchers have proposed a variety of mitigation strategies~\cite{wan2024surveybiastexttoimagegeneration,t2ibiassurvey}. 
These interventions span the entire content generation pipeline, from prompt engineering techniques that manually add diversity cues~\cite{luo2024versusdebias}, to computationally expensive model fine-tuning on balanced datasets~\cite{shen2023finetuning}, and diffusion process manipulation that adjusts the sampling trajectory at inference time~\cite{parihar2024balancing}. 
While effective to varying degrees, these methods can be inflexible, computationally intensive, or fail to address the underlying biased associations within the model's core components.

\begin{figure}
    \centering
    \includegraphics[width=\linewidth]{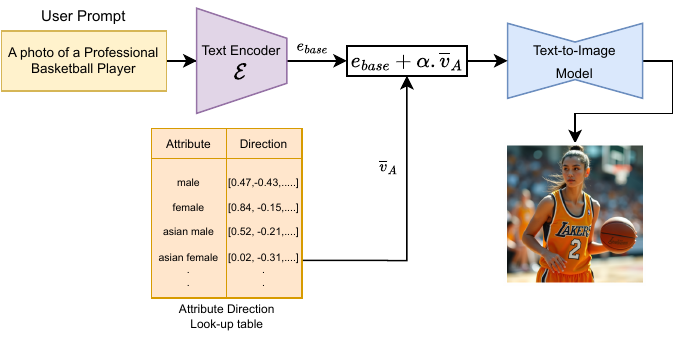}
    \caption{Diverse Image Generation using Embedding Arithmetic $EA$. During inference, a user's prompt is converted into base embedding $e_{base}$ using a text encoder $\mathcal{E}$. A random pre-calculated direction vector, $\overline{v}_{\mathcal{A}}$, is retrieved from the look-up table and added to the base embedding. The strength of the attribute is controlled by a scaling factor $\alpha$. This new, modified embedding is then passed to the T2I model to produce the final output.}
    \label{fig:process}
\end{figure}

\begin{figure*}
    \centering
    \includegraphics[width=\linewidth]{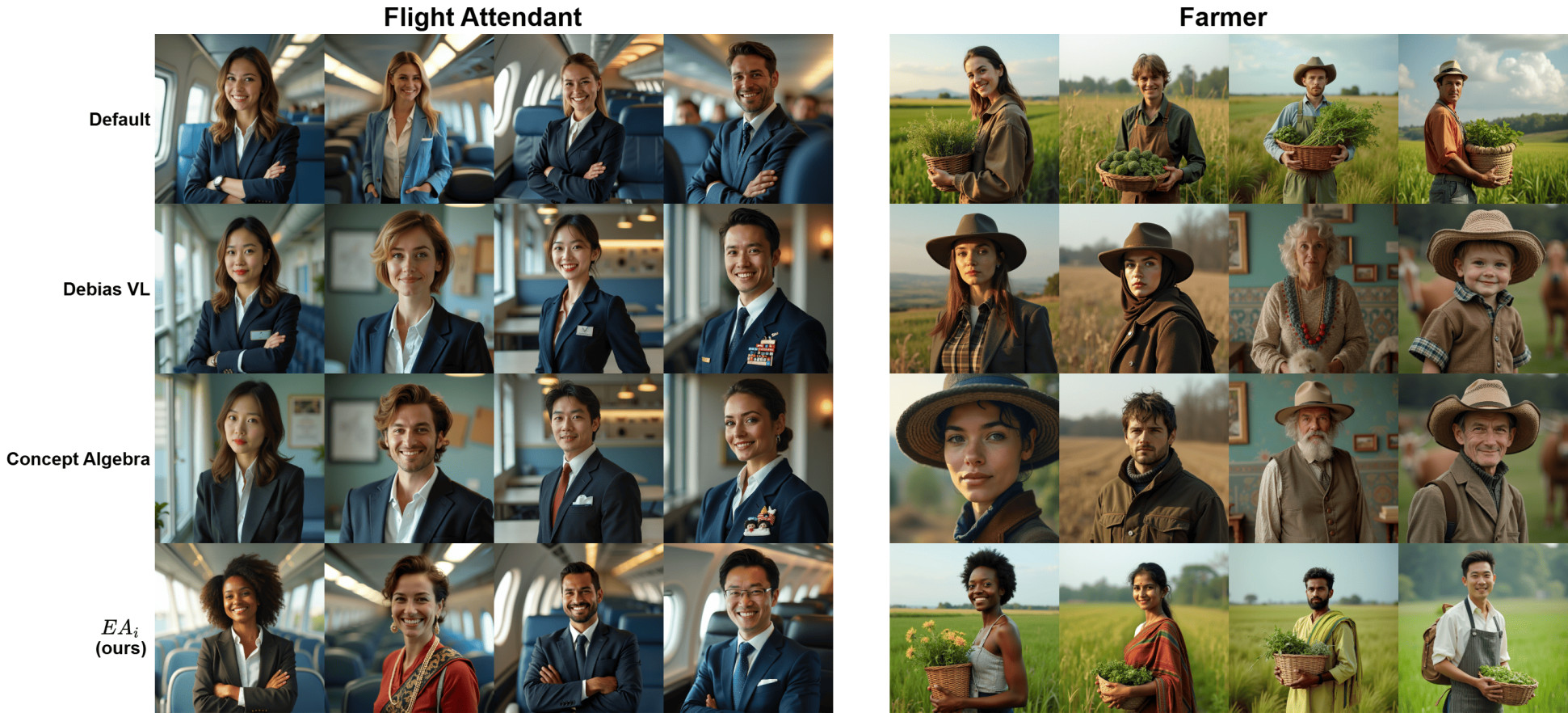}
    \caption{Qualitative comparison showing our method ($EA_i$) improves intersectional diversity while preserving context (e.g., aircraft background).}
    \label{fig:qualitative_benchmark}
\end{figure*}

One class of methods that has attracted significant research attention involves manipulating conditional embeddings to achieve more diverse demographic representations in generated images~\cite{chuang2023debiasing,wang2023concept,kim2025rethinking,hirotasaner,bonna2024debiaspi}. 
Since these approaches typically do not require specially curated datasets or impose substantial computational overhead during generation, they are particularly effective for practical applications. 
Within this class, some methods optimize a lightweight layer following the text encoder~\cite{hirotasaner}, while others directly augment the text embeddings to promote diversity in image generation~\cite{chuang2023debiasing,kim2025rethinking,wang2023concept}.

We adapt the geometric intuition of directional concept vectors from NLP~\cite{bolukbasi2016man,mikolov2013efficient} to the text embeddings of T2I models. 
We address a critical gap in the literature by systematically evaluating this approach's ability to map the geometry of bias and effectively mitigate it in state-of-the-art representations.

In this work, we adapt and analyze Embedding Arithmetic as a lightweight, model-agnostic framework that serves as both a scientific probe to map the geometry of bias and as a practical tool for its correction. 
We move beyond simply applying this technique and conduct a rigorous, hypothesis-driven investigation into the latent space of the conditional text encoder of state-of-the-art T2I models. Our primary contributions are: (1) We provide a characterization of the latent geometry in the text encoders used in state-of-the-art T2I models, demonstrating its entangled, approximately compositional, and locally linear properties. (2) Our method significantly outperforms similar baselines in improving gender and racial diversity for unseen concepts, offering a superior solution to the critical fairness-coherence trade-off by uniquely maintaining high concept coherence. (3) Our lightweight approach relies solely on the learned knowledge of the T2I model and requires no fine-tuning or specialized datasets.

\section{Background and Related Works}

Bias mitigation in generative models has been approached from multiple angles, each targeting a different stage of the generation pipeline. 
Many methods intervene at inference time, either through prompt engineering that manually adds diversity cues to the input text~\cite{bonna2024debiaspi,cheng2024conditional}, or by directly manipulating the diffusion process to adjust the noise sampling trajectory towards a fairer target distribution~\cite{parihar2024balancing,kang2025fairgen}. 
A more direct approach involves altering the model's parameters to address bias at its source, either through full retraining on balanced datasets or, more commonly, through parameter-efficient fine-tuning methods like LoRA~\cite{hu2022lora} to steer demographics in the generated images~\cite{gandikota2024concept,shen2023finetuning}. 
While effective, these methods can be computationally intensive, place the debiasing burden on the user, or lack fine-grained control.

\paragraph{Conditional Embedding Manipulation} These methods focus on altering the text embeddings that condition the image generation process. 
Some methods learn additive transformations or specialized tokens to augment the prompt embedding and steer the model towards producing diverse outputs~\cite{Zhang_2023_ICCV,kim2023stereotyping,hirotasaner}.

A key line of research within this category operates on the principle that biases are geometrically encoded as directions or subspaces in the embedding space.
This geometric view was pioneered in NLP by Bolukbasi et al.~\cite{bolukbasi2016man}.
They demonstrated that gender stereotypes could be represented by a vector (e.g., the difference vector between embeddings for "he" and "she").
They proposed mitigating bias by projecting word embeddings onto a subspace orthogonal to this bias direction, effectively "nulling out" the stereotypical association.
This concept has since been applied to other domains, including image retrieval~\cite{couairon2022embedding} and controlling AI assistants~\cite{chen2025personavectorsmonitoringcontrolling}.
In the context of image generation, Radford et al.,~\cite{radford2015unsupervised} performed similar vector arithmetic on the conditional embeddings of GAN models to alter various properties of the generated images.
Brack et al.,~\cite{NEURIPS2023_4ff83037} performs similar vector arithmetic on the conditional embedding space to perform image editing.

This foundational geometric approach has also been adapted as a method for debiasing T2I models. 
For instance, DebiasVL \cite{chuang2023debiasing} learns a linear projection matrix that aims to equalize the embeddings of concept pairs with different sensitive attributes (e.g., "male doctor" and "female doctor"). 
Concept Algebra \cite{wang2023concept}, while sharing the intuition of vector operations, identifies concept directions within the noise space of the diffusion model's U-Net and performs manipulations during the denoising steps.

Our work returns to this original, simpler idea of direct vector arithmetic in the text embedding space, similar in spirit to the original work in NLP. 
However, our contribution is twofold: First, unlike methods that require optimizing a projection matrix \cite{chuang2023debiasing} or intervening in the complex denoising process \cite{wang2023concept}, we show that simple, pre-computed attribute vectors are remarkably effective for debiasing state-of-the-art transformer-based T2I models. 
Second, we use this simple arithmetic not just as a bias mitigation tool, but also as an interpretable probe to conduct a rigorous, hypothesis-driven analysis of the latent space's geometry, providing novel insights into its entangled and locally linear structure.

\section{Methods}
We propose Embedding Arithmetic, a framework for isolating and composing semantic concepts as vectors. 
Using this method, we evaluate three key hypotheses about the latent space of state-of-the-art T2I models: disentanglement, composability, and local linearity.

\begin{figure}[h]
    \includegraphics[width=\linewidth]{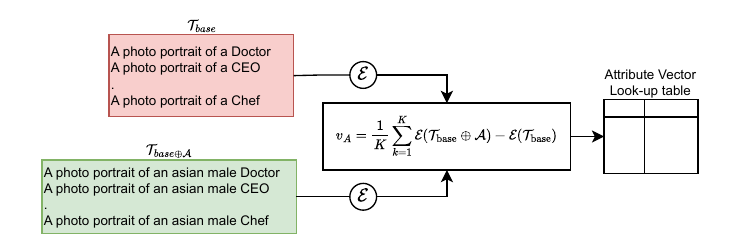}
    \caption{Creating an Attribute Direction Vector. The vector $v_\mathcal{A}$ representing the direction of a specific attribute is calculated by finding the average difference between the text embeddings of prompts containing the attribute ($\mathcal{T}_{base \oplus A}$) and those without it ($\mathcal{T}_{base}$). This pre-calculated vector is then stored in a look-up table to be used for steering the image generation.}
    \label{fig:make_look_up}
\end{figure}

\subsection{Problem Formulation}
\label{ssec:isolate}
Let $\mathcal{E}:\mathcal{T} \xrightarrow[]{} \mathbb{R}^d$ be a text encoder that maps a text prompt $\mathcal{T}$ into $d$ dimensional embedding space.
Our framework operates on these embeddings.

\paragraph{Isolating and generalizing attribute Vectors :}
An attribute vector for a specific attribute $\mathcal{A}$ is first isolated by taking the vector difference between the embeddings generated from a base prompt $\mathcal{T}_k$ and the same prompt with the attribute, $\mathcal{T}_k \oplus \mathcal{A}$ (see Figure \ref{fig:make_look_up}).
Similar to Radfort et al., \cite{radford2015unsupervised}, we average this difference vector across $K$ set of  diverse base concepts (professions), in-order to create a robust, context-independent representation:
$v_{\mathcal{A}} = \frac{1}{K} \sum_{k=1}^{K}\mathcal{E}(\mathcal{T}_{\text{base}} \oplus \mathcal{A}) - \mathcal{E}(\mathcal{T}_{\text{base}})$. 
This averaged vector $v_{A}$ represents the generalized direction for $\mathcal{A}$.
To ensure it represents a pure direction, independent of magnitude, we normalize it to a unit vector for our analysis: $v_{\mathcal{A}} = \frac{v_{\mathcal{A}}}{\left\lVert v_{\mathcal{A}} \right\rVert_2}$.

\paragraph{Composition :}
The final target embedding $e_{target}$ used to condition the generative model is composed via a linear combination of a base embedding $e_{base}$ and a set of $N$ attribute vectors (see Figure \ref{fig:process}):
$e_{target} = e_{base} + \sum_{i=1}^{N} \alpha_{i} . v_{\mathcal{A}_{i}}$.
where $\alpha_{i}$ is a scalar weight controlling the steering strength for $i$-th attribute.

\subsection{Experimental Setup}
\label{ssec:setup}
\paragraph{Text-to-Image Generator:} For image synthesis, we employ the FLUX 1.0-Dev model\footnote{\url{https://huggingface.co/black-forest-labs/FLUX.1-dev}}. 
Its core architecture is a Multi-Modal Diffusion Transformer (MM-DiT) \cite{esser2024scaling}, which is conditioned using the pooled text embeddings ($[CLS]$ token)  from the OpenAI CLIP ViT-L/14 text encoder \cite{radford2021learning}.
For all our experiments we set the denoising parameters as recommended by the developers of the model, the guidance scale is set to 3.5, number of denoising steps to 26, and an output image size of $512 \times 512$.

\paragraph{Text Encoder ($\mathcal{E}$):} Although FLUX 1.0-Dev utilizes both CLIP~\cite{radford2021learning} and T5~\cite{2020t5}, we strategically target the CLIP embedding space. 
We base this on the distinct roles of these encoders: T5 sequence embeddings govern fine-grained linguistic structure, whereas CLIP pooled embeddings encapsulate global semantic attributes, including demographics. 
Since societal biases manifest as high-level semantic associations, manipulating the global CLIP vector is sufficient for effective steering. This design aligns with prior bias mitigation works~\cite{hirotasaner,chuang2023debiasing,li2024self} and also significantly improves computational efficiency.

\paragraph{Concepts and Attributes:} To evaluate the robustness of our framework, we selected a diverse set of base concepts representing professions. 
Specifically, we employed ten distinct occupations: "Interior Designer", "Doctor", "Engineer", "Teacher", "Childcare Worker", "Clergy", "Police Officer", "CEO", "Artist", and "Mechanic". 
These base concepts were randomly sampled from the 146 professions reported by the U.S. Bureau of Labor Statistics (BLS), adhering to the experimental setup established in \cite{luccioni2023stable}.

For the demographic attributes, we focused on two primary axes: Gender ("male", "female") and Race ("white", "black", "asian", "indian"). 
Our analysis is restricted to binary gender categories because datasets and pretrained attribute detectors for non-binary genders are currently unavailable~\cite{wan2024surveybiastexttoimagegeneration}. 
Furthermore, the selected racial categories were chosen based on the empirically demonstrated reliability of our attribute predictor for these specific groups, as detailed in Appendix \ref{sec:fairface_performance}.

\paragraph{Look up Table:} Our lookup table contains 14 averaged attribute vectors ($v_{\mathcal{A}}$) for the gender and race attributes individually as well as their intersections.
These vector are calculated as described in Section \ref{ssec:isolate}, using the base concepts.
This pre-computation allows for lightweight, inference-time steering without needing to calculate vectors on the fly.
During inference, we sample attribute vectors based on preferred distribution and then compose the attribute vector to the embeddings of the base prompt ($e_{base})$.

\paragraph{Attribute Detectors:} We utilize a ResNet~\cite{he2016deep} based model trained on the FairFace dataset \cite{karkkainenfairface} to identify gender and race attributes in the generated images. 
We validated the detector on a curated synthetic dataset, achieving $>$90\% accuracy (see Appendix \ref{sec:fairface_performance} for metrics).

\subsection{Experimental Design for Probing Latent Space Properties}
To investigate the geometric and functional properties of the latent space, we designed four experiments, each testing a specific hypothesis.

\paragraph{\textbf{Hypothesis 1 (Orthogonality)}:} \textit{If semantic concepts are learned in a disentangled manner, their corresponding directional vectors should be approximately orthogonal in the embedding space.}

We formally test this hypothesis by computing the pairwise cosine similarity between all generalized attribute vectors $\{\hat{v}_{\mathcal{A}}\}$. 
For any two vectors $\hat{v}_{\mathcal{A}_i}$ and $\hat{v}_{\mathcal{A}_j}$ representing conceptually distinct attributes, their similarity is calculated as:
$\textit{similarity}(\hat{v}_{\mathcal{A}_i}, \hat{v}_{\mathcal{A}_j}) = \frac{\hat{v}_{\mathcal{A}_i} \cdot \hat{v}_{\mathcal{A}_j}}{\|\hat{v}_{\mathcal{A}_i}\| \|\hat{v}_{\mathcal{A}_j}\|}$.
A value close to 0 indicates orthogonality and provides evidence for disentanglement. We visualize these relationships in a similarity matrix.

\paragraph{\textbf{Hypothesis 2 (Composability)}:} \textit{    If the latent space is perfectly compositional, the vector for an intersectional concept should be equivalent to the linear sum of its constituent attribute vectors.}

We test this by comparing two vector types for an intersectional attribute (e.g., "black female"): the Composed Vector ($\hat{v}_{\text{composed}} = \hat{v}_{\text{black}} + \hat{v}_{\text{female}}$) and the Intersectional Vector ($\hat{v}_{\text{black} \oplus \text{female}}$).
The comparison is two-fold: 1) we calculate their direct cosine similarity, where a value less than 1.0 indicates imperfect compositionality, and 2) we analyze and compare the demographic distributions of images generated by the T2I when conditioned on base concept embeddings composed using each vectors.

\paragraph{\textbf{Hypothesis 3 (Local Linearity)}:} \textit{The semantic manifold is locally linear, meaning that vector translations are effective and predictable withing a bounded region around the base concept.}

We test this by defining the target embedding as a function of steering strength $\alpha$ and sweeping its value across the range $[0, 5.0]$. We quantify the effect by measuring the attribute classifier's confidence score and the newly introduced concept coherence score (see Section~\ref{ssec:concept_consistency}). 
A monotonic relationship between $\alpha$ and the attribute confidence supports the hypothesis, while a drop in concept coherence identifies the boundary of the linear region where concept drift occurs.

\paragraph{\textbf{Hypothesis 4 (Generalizability)}:} \textit{The generalized attribute vectors ($v_\mathcal{A}$) are not overfitted to the source concepts used for deriving them and should generalize to a disjoint set of unseen concepts.}

To test for generalization, we partition our concepts into a $\mathcal{P}_{\text{source}}$ set and a held-out $\mathcal{P}_{\text{target}}$ set. We compute the generalized semantic vectors $v_\mathcal{A}$ using only $\mathcal{P}_{\text{source}}$. We then apply these vectors to the base concepts in $\mathcal{P}_{\text{target}}$:
$\vec{e}_{\text{target}} = \mathcal{E}(\mathcal{T}_{\text{target}}) + \alpha \cdot v_\mathcal{A}, \quad \text{where} \quad \mathcal{T}_{\text{target}} \in \mathcal{P}_{\text{target}}$.
A successful transfer is indicated by a statistically significant shift in the demographic distribution of the generated images, without a corresponding degradation of the coherence to the base concept (see Section \ref{ssec:quant}).

\subsection{Baselines}
\label{ssec:baselines}
We benchmark our framework against two established training-free methods for bias reduction in Text-to-Image models: DebiasVL~\cite{chuang2023debiasing} and Concept Algebra~\cite{wang2023concept}.
To ensure a rigorous comparison on state-of-the-art technology, we re-implemented both methods on the modern rectified flow architectures of FLUX 1.0-Dev and Stable Diffusion 3.5-Large \cite{esser2024scaling}.
By modifying DebiasVL's projection matrices and Concept Algebra's noise-space interventions, we successfully adapted both frameworks to support precise intersectional steering tasks.

\subsection{Evaluation Metrics}
\label{ssec:metrics}
Defining unique attribute distributional requirements for individual concepts like professions is subjective and non-trivial.
Therefore, following similar works \cite{kim2025rethinking,luccioni2023stable,friedrich2025auditing}, our objective is to ensure that all attributes are uniformly distributed across the final sets of generated images.
We validate our method against baselines based on two parameters, 1) The amount of demographic diversity on the images generated by different methods on different general base prompts, 2) The faithfulness of the method in maintaining the properties of the base prompts.

\subsubsection{Shannon Entropy for Attribute Diversity: } \ Several previous work relied on the ratio of the minority attribute for a specific profession to monitor the demographic diversity on the generated images \cite{kim2025rethinking,chuang2023debiasing,hirotasaner}.
Since prior work uses metrics limited to binary categories (e.g., male/female), our inclusion of multi-category race attributes necessitates a more general diversity measure. 
We therefore employ Shannon entropy: $H(X) = - \sum_{i=1}^{k} p(x_i) \, \log_{N} p(x_i)$.
where $X$ is the overall distribution across all the attributes, $N$ is set to 2 for gender attributes and 4 for race attributes.
$H(X)$ tends to 1 if the categories are equally distribution, while it tends to 0 if the distribution is completely skewed to one specific category.

\subsubsection{Concept Coherence Score (CCS): } \
\label{ssec:concept_consistency}
To evaluate a model's faithfulness in preserving the base concept, we introduce the Concept Coherence Score (CCS). 
To avoid the circularity of using CLIP Score~\cite{hessel-etal-2021-clipscore} (since CLIP encodes the prompts), we ensure independence by adapting the VQA-based approach similar to~\cite{luccioni2023stable}.
Our CCS is derived by prompting an ensemble of diverse VQA models with the question, \textit{“Answer only with yes or no: Is the person in this image a \{concept\}?”}, where \textit{\{concept\}} depicts a gender neutral profession term like "Carpenter". 
We then compute the mean softmax probability for the token \texttt{yes} across the ensemble. This provides a continuous, fine-grained measure of confidence that the generated image reflects the intended base concept, with higher values indicating stronger coherence. 
Further implementation details are provided in the Appendix \ref{app:ccs_details}.

\section{Results}
Before detailing our geometric analysis, we established the robustness of our attribute vector construction. All subsequent experiments utilize a reference set size of $K=10$. 
This choice is grounded in our ablation study (detailed in Appendix~\ref{sec:app_k_ablation}), which demonstrates that averaging across ten diverse professions effectively diversify the demographics in the target neutral concepts than using a single reference ($K=1$), while also maintaining the concept coherence.

\begin{figure*}[t]
    \centering
    \includegraphics[width=\linewidth]{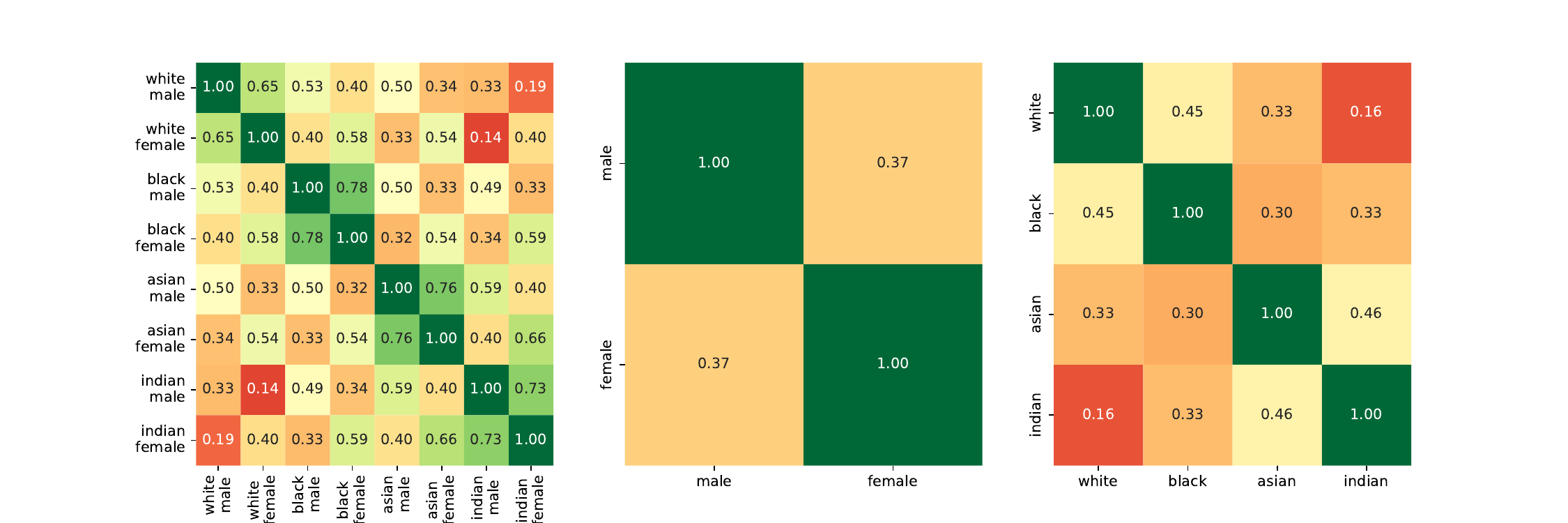}
    \caption{ Pairwise cosine similarity heatmaps for (left) intersectional race and gender vectors, (center) generalized gender vectors, and (right) generalized race vectors}
    \label{fig:hypothesis1}
\end{figure*}

\begin{figure}[h]
    \centering
    \begin{subfigure}[t]{0.49\linewidth}
        \centering
        \includegraphics[width=\linewidth]{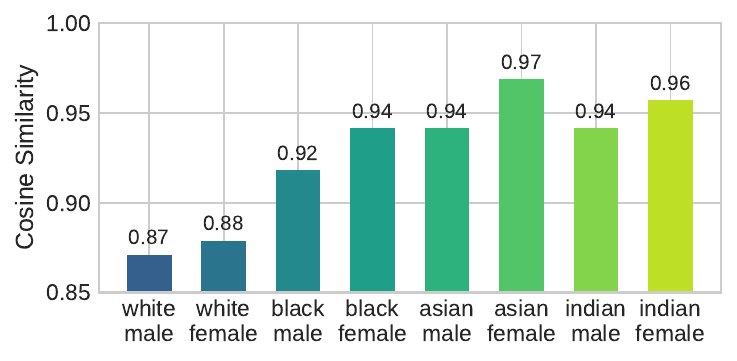}
        \caption{}
        \label{fig:compositionality}
    \end{subfigure}
    \hfill
    \begin{subfigure}[t]{0.49\linewidth}
        \centering
        \includegraphics[width=\linewidth]{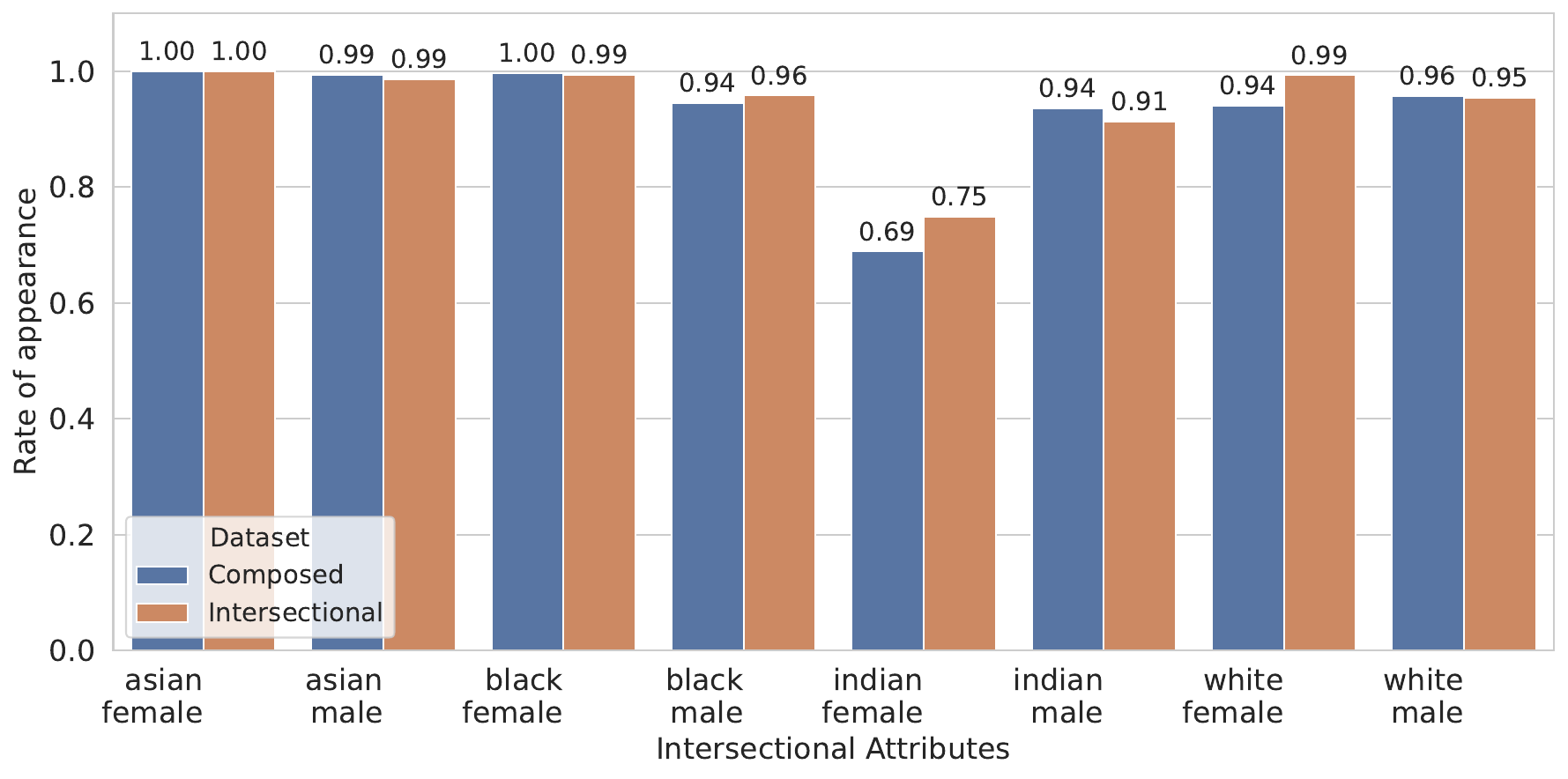}
        \caption{}
        \label{fig:compositionality2}
    \end{subfigure}
    \caption{(a) The high similarity across all categories confirms that linear composition is a strong but imperfect approximation of intersectionality in the latent space. (b) Attribute generation rates for intersection of "race" and "gender" concepts when conditioning on a Composed Vector versus an  Intersectional Vector (N=100 images per condition).}
\end{figure}

\subsection{Geometric Properties: Entanglement and Composability}
Testing Hypothesis 1 (Figure \ref{fig:hypothesis1}) refutes the notion of orthogonality, instead revealing significant entanglement. 
Generalized vectors for "Male" and "Female" exhibit positive correlation, suggesting they share underlying variance (e.g., "human subject") rather than acting as pure opposites. 
Similarly, race vectors form correlated clusters, confirming the latent space is a complex, non-orthogonal manifold.

Testing Hypothesis 2 (Figure \ref{fig:compositionality}) confirms that while linear composition (e.g., $v_{white} + v_{female}$) is a strong approximation ($>0.87$ cosine similarity to intersectional vectors), it is not perfect. Practically, however, this deviation is minor: demographic generation rates using Composed vs. Intersectional vectors differ by $<3\%$ for most identities (Figure \ref{fig:compositionality2}). Exceptions, such as the lower accuracy for "Indian Female," likely stem from data under-representation or known detector biases (see Appendix \ref{sec:fairface_performance}), highlighting the dependency of fairness metrics on detector robustness.

\begin{figure}[h]
    \centering
    \begin{subfigure}[t]{0.49\linewidth}
        \centering
        \includegraphics[width=\linewidth]{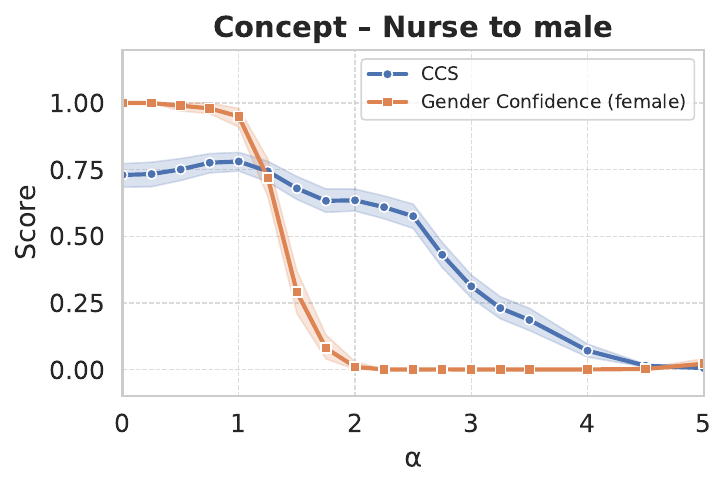}
        \caption{}
        \label{fig:h3_nurse}
    \end{subfigure}
    \hfill
    \begin{subfigure}[t]{0.49\linewidth}
        \centering
        \includegraphics[width=\linewidth]{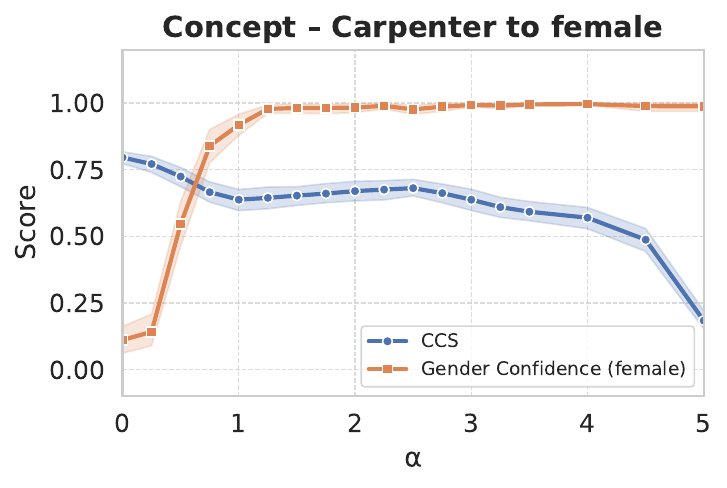}
        \caption{}
        \label{fig:h3_carpenter}
    \end{subfigure}
    
    \caption{Probing for local linearity by steering (a) "Nurse" towards "male" and (b) "Carpenter" towards "female". The orange line (female Gender Confidence) shows a clear monotonic response to steering strength $\alpha$. The blue line (Concept Coherence Score) remains stable within a bounded region.}
    \label{fig:consistency_gender}
\end{figure}

\begin{figure*}[h]
    \centering
    \begin{subfigure}[t]{\linewidth}
        \centering
        \includegraphics[width=\linewidth]{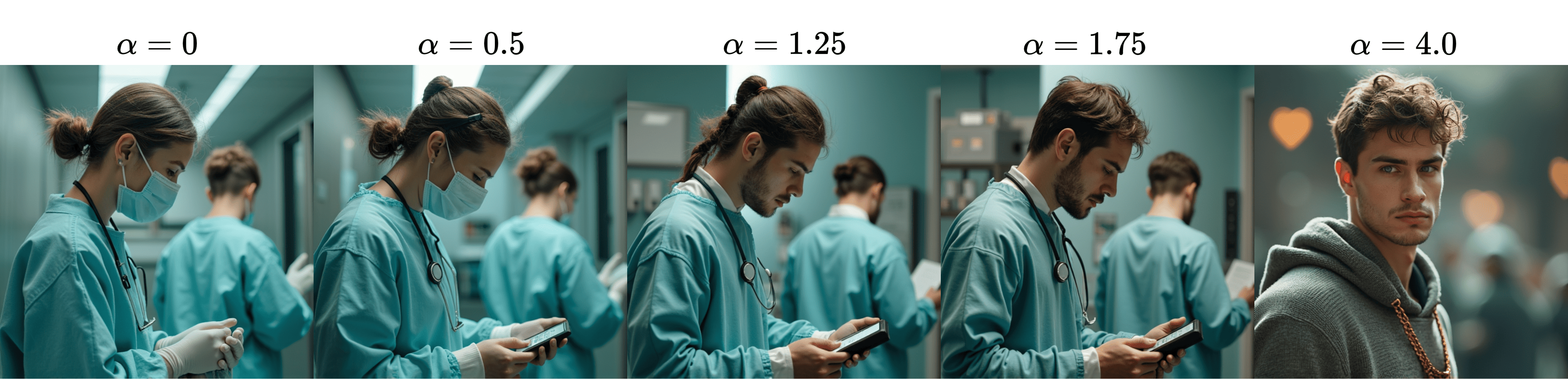}
        \caption{Nurse $\to$ Male}
        \label{fig:h3_nurse_qualitative}
    \end{subfigure}
    \vfill
    \begin{subfigure}[t]{\linewidth}
        \centering
        \includegraphics[width=\linewidth]{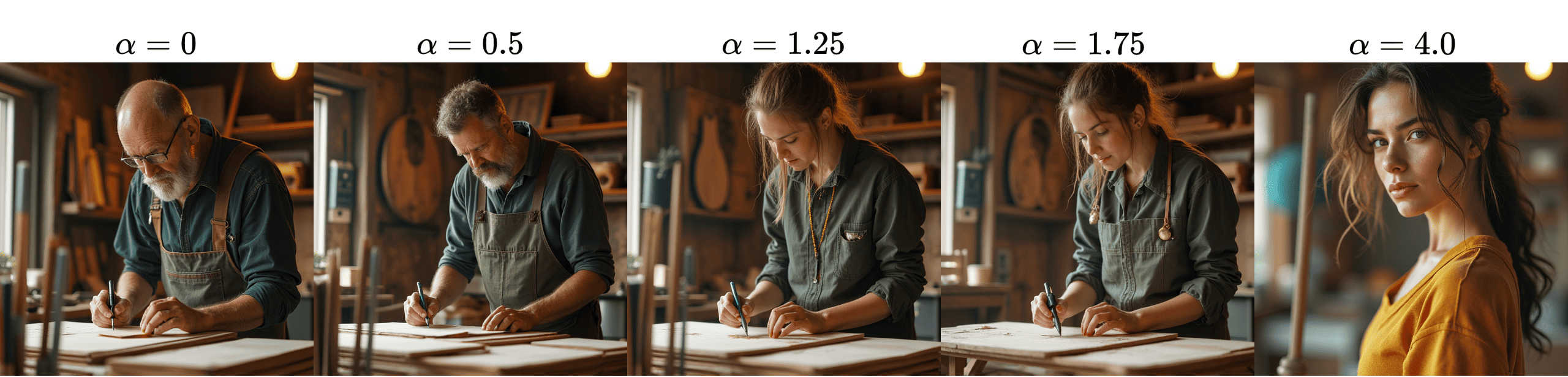}
        \caption{Carpenter $\to$ Female}
        \label{fig:h3_carpenter_qualitative}
    \end{subfigure}
    
    \caption{Qualitative examples representing the effect on steering strength $\alpha$. Increasing the $\alpha$ smoothly transitions the gender of the Nurse (a) and Carpenter (b) until the concept coherence breaks down at extreme value ($\alpha=4.0$)}
    \label{fig:alpha_consistency_gender}
\end{figure*}

\begin{table*}[t]
\caption{Quantitative results for the Generalization Hypothesis on four unseen professions, using the FLUX.1-dev model. We report diversity using gender entropy ($H_g$) and race entropy ($H_r$), and concept integrity using the Concept Coherence Score (CCS). Higher is better for all metrics. Our Embedding Arithmetic ($EA$) methods, particularly the method using intersectional attribute vectors ($EA_i$), achieve the best diversity scores while maintaining high concept coherence. The best result in each column is in \textbf{bold}, and the second best is \underline{underlined}.}
\label{tab:my-table}
\resizebox{\textwidth}{!}{%
\begin{tabular}{clcccccccccccc}
\hline
\multicolumn{1}{l}{}     &                     & \multicolumn{3}{c}{Fire Fighter}                                                                            & \multicolumn{3}{c}{Mechanic}                                                                                & \multicolumn{3}{c}{Flight Attendant}                                                                        & \multicolumn{3}{c}{Nurse}                                                                                   \\ \hline
\multicolumn{1}{l}{}     &                     & $H_g$ & $H_r$ & CCS & $H_g$ & $H_r$ & CCS & $H_g$ & $H_r$ & CCS & $H_g$ & $H_r$ & CCS \\ \hline
\multirow{5}{*}{Flux}    & Default             & 0.50                                 & 0.17                               & 0.58                            & 0.08                                 & 0.08                               & \textbf{0.64}                   & 0.41                                 & 0.18                               & \textbf{0.62}                   & 0.00                                 & 0.21                               & \underline{0.57}                      \\
                         & Debias VL           & 0.88                                 & 0.23                               & 0.60                            & 0.86                                 & 0.50                               & 0.52                            & 0.70                                 & 0.34                               & 0.60                            & 0.00                                 & 0.24                               & \textbf{0.58}                   \\
                         & Concept Algebra     & 0.97                                 & 0.16                               & \underline{0.61}                      & 0.91                                 & \underline{0.28}                         & 0.55                            & 0.24                                 & 0.33                               & 0.59                            & 0.66                                 & 0.30                               & 0.57                            \\
                         & $EA_g$         & \textbf{1.00}                        & 0.08                               & 0.58                            & \textbf{1.00}                        & 0.22                               & \underline{0.61}                      & \underline{0.99}                           & 0.22                               & 0.61                            & \textbf{0.95}                        & 0.23                               & 0.57                            \\
                         & $EA_i$ & \underline{0.99}                           & \textbf{0.82}                      & \textbf{0.61}                   & \underline{0.99}                           & \textbf{0.93}                      & 0.60                            & \textbf{0.99}                        & \textbf{0.79}                      & \underline{0.62}                      & \underline{0.68}                           & \textbf{0.80}                      & 0.57                            \\ \hline
\multirow{5}{*}{SD3.5-large} & Default             & 0.10                                 & 0.47                               & 0.61                            & 0.00                                 & 0.14                               & \textbf{0.69}                   & 0.00                                 & 0.38                               & \underline{0.67}                      & 0.00                                 & 0.16                               & \textbf{0.62}                   \\
                         & Debias VL           & 0.61                                 & \underline{0.48}                         & \textbf{0.62}                   & 0.38                                 & 0.19                               & \underline{0.68}                      & 0.20                                 & \underline{0.61}                         & 0.64                            & 0.21                                 & 0.24                               & 0.55                            \\
                         & Concept Algebra     & 0.73                                 & 0.35                               & 0.58                            & 0.65                                 & \underline{0.35}                         & 0.54                            & \textbf{0.67}                        & 0.55                               & 0.61                            & 0.68                                 & \underline{0.39}                         & 0.48                            \\
                         & $EA_g$         & \textbf{0.99}                        & 0.36                               & 0.62                            & \textbf{0.99}                        & 0.07                               & 0.66                            & 0.39                                 & 0.37                               & \textbf{0.68}                   & {\ul 0.81}                           & 0.16                               & 0.60                            \\
                         & $EA_i$ & \underline{0.99}                           & \textbf{0.91}                      & \underline{0.62}                      & \underline{0.99}                           & \textbf{0.96}                      & 0.61                            & {\ul 0.53}                           & \textbf{0.94}                      & 0.66                            & \textbf{0.85}                        & \textbf{0.98}                      & \underline{0.61}                      \\ \hline
\end{tabular}%
}
\end{table*}

\subsection{The Latent Space is Locally Linear}
\label{ssec:hyp3_results}
To test the Local Linearity Hypothesis (Hypothesis 3), we applied gender-steering vectors to "Carpenter" (male-biased) and "Nurse" (female-biased) by sweeping the steering strength $\alpha$.
These two concepts exhibit extreme cases of demographic bias towards the gender direction.
They are also not used to compute the attribute vectors in our look-up table.
The results (Figure \ref{fig:consistency_gender}) demonstrate a predictable, monotonic relationship between $\alpha$ and the target gender confidence. 
The gender for "Carpenter" reliably inverts to female at $\alpha \approx 0.75$, while the more strongly biased "Nurse" inverts to male at $\alpha \approx 1.5$. The Concept Coherence Score (CCS) quantitatively defines the bounded region of this effect, remaining robust through the gender transition before declining sharply at extreme steering strengths ($\alpha >  3.0$). Qualitative examples in Figure \ref{fig:alpha_consistency_gender} visually confirm this boundary, empirically proving that the semantic manifold is locally linear within a predictable range.

\subsection{Quantitative Analysis of Generalization to Unseen Concepts}
\label{ssec:quant}
To test the Generalization Hypothesis (Hypothesis 4), we applied our Embedding Arithmetic ($EA$) framework to four unseen professions—chosen for their strong gender skew in the LAION-5B dataset \cite{seshadri-etal-2024-bias} and  compared its performance against the default models and two baselines. 
We tested both gender only ($EA_g$) and intersectional ($EA_i$) attribute vectors, with a fixed steering strength of $\alpha=1.25$. 
For each profession, we generated 100 images using the prompt template \texttt{A photo portrait of \{article\} \{profession\}.} and evaluated them on our key metrics. 
The full results for both FLUX.1-Dev and Stable Diffusion 3.5-Large are in Table \ref{tab:my-table}, with a qualitative comparison in Figure \ref{fig:qualitative_benchmark}.

\paragraph{\textbf{Analysing Baseline Bias}}
The default generations from both models confirm that the unseen professions exhibit strong stereotypical biases.
For concepts like "Mechanic" and "Nurse," gender diversity is extremely low ($H_g \le 0.08$), indicating the models almost exclusively generate one gender. 
Similarly, racial diversity ($H_r$) is poor across all professions, consistently scoring below 0.22. 
This establishes a clear baseline of stereotypical output that our method aims to correct. 
The FLUX model shows slightly better gender diversity for "Fire Fighter" and "Flight Attendant," suggesting its training data may have had better representation for these concepts, though this is also impossible to verify as the dataset is proprietary.

\paragraph{\textbf{Comparative Performance}}
Our results highlight the critical challenge of achieving comprehensive fairness without sacrificing image quality. 
While all intervention methods improve gender representation, they often compromise concept coherence (e.g., for "Mechanic," CCS drops from 0.64 to 0.55 with DebiasVL), a known trade-off \cite{kim2025rethinking}. 
Concept Algebra proves inconsistent; while effective for male-biased professions like "Mechanic" $(H_g=0.91$), it fails on the female-biased "Flight Attendant" reducing gender diversity ($H_g$ drops from 0.41 to 0.24).
Furthermore, its impact on racial diversity is marginal at best.

In contrast, our Embedding Arithmetic ($EA$) framework demonstrates superior control and efficacy. 
The $EA_i$ variant outperforms all other methods by simultaneously improving both gender and racial diversity, a significant advantage over the baselines, while maintaining high concept coherence. 
Crucially, these findings are not specific to a single T2I architecture. 
As shown in Table \ref{tab:my-table}, we replicated our experiments on the Stable Diffusion 3.5-Large model and observed the same pattern: our framework consistently and effectively mitigated the model's strong stereotypical biases, proving its general applicability as a tool for steering T2I models towards fairer outcomes.

The robustness of our framework is further supported by additional qualitative results in the Appendix \ref{app:Further_results} and \ref{app:beyond_demographics}, where we also showcase its broader utility by applying it to non-demographic concepts like artistic style and expressions.

In summary, these results validate our Generalization Hypothesis (Hypothesis 4), proving that our generalized attribute vectors transfer effectively to unseen concepts and that our framework provides a more controllable means of mitigating bias than existing methods.

\section{Limitations}
While Embedding Arithmetic offers a lightweight and effective path to fairness, it is not without limitations.
First, our method operates on the global pooled embedding of the CLIP encoder. 
Consequently, the steering effect is applied uniformly across the generated context. 
For group prompts (e.g., "a team of doctors"), this could cause demographic homogenization rather than diverse group representation.
Second, by bypassing the T5-sequence embeddings, we sacrifice fine-grained, token-level control. 
This limits the framework's ability to handle attribute binding in complex, multi-subject prompts (e.g., correctly assigning different genders to two distinct subjects in the same scene).
Finally, our evaluation relies on external attribute classifiers (e.g., FairFace) to quantify success. 
As noted in our analysis of racial categories (refer Appendix \ref{sec:fairface_performance}), these classifiers possess inherent biases and limitations in generalization, potentially skewing fairness metrics for underrepresented groups. 
Future work should explore hybrid approaches that combine global embedding arithmetic with spatial attention control to address these granularity issues.

\section{Conclusion}
In this work, we established Embedding Arithmetic as both a rigorous scientific probe and a practical intervention tool for text-to-image fairness.
Through a hypothesis-driven analysis of state-of-the-art models like FLUX 1.0-Dev, we characterized the conditional embedding space as a globally entangled yet locally linear manifold. 
Leveraging this geometric insight, we introduced an intersectional steering method that significantly outperforms existing baselines in improving gender and racial diversity for unseen concepts. 
Crucially, our framework achieves this without fine-tuning and with minimal computational overhead, uniquely resolving the tension between promoting fairness and preserving the visual coherence of the base concept.
Ultimately, while inference-time interventions serve as powerful, immediate correctives, they are not a panacea; achieving comprehensive fairness will likely require integrating such geometric insights directly into the training objectives of future generative models.

\subsubsection{Acknowledgements:} This project is funded by the Bayerische Transformations- und Forschungsstiftung under the project EvenFAIr (AZ-1611-23).

\bibliographystyle{splncs04}
\bibliography{main}

\begin{thebibliography}{10}
\providecommand{\url}[1]{\texttt{#1}}
\providecommand{\urlprefix}{URL }
\providecommand{\doi}[1]{https://doi.org/#1}

\bibitem{baldridge2024imagen}
Baldridge, J., Bauer, J., Bhutani, M., Brichtova, N., Bunner, A., Chan, K., Chen, Y., Dieleman, S., Du, Y., Eaton-Rosen, Z., et~al.: Imagen 3. CoRR  (2024)

\bibitem{bolukbasi2016man}
Bolukbasi, T., Chang, K.W., Zou, J.Y., Saligrama, V., Kalai, A.T.: Man is to computer programmer as woman is to homemaker? debiasing word embeddings. Advances in neural information processing systems  \textbf{29} (2016)

\bibitem{bonna2024debiaspi}
Bonna, S., Huang, Y.C., Novozhilova, E., Paik, S., Shan, Z., Feng, M.Y., Gao, G., Tayal, Y., Kulkarni, R., Yu, J., et~al.: Debiaspi: inference-time debiasing by prompt iteration of a text-to-image generative model. In: European Conference on Computer Vision. pp. 68--83. Springer (2024)

\bibitem{NEURIPS2023_4ff83037}
Brack, M., Friedrich, F., Hintersdorf, D., Struppek, L., Schramowski, P., Kersting, K.: Sega: Instructing text-to-image models using semantic guidance. In: Oh, A., Naumann, T., Globerson, A., Saenko, K., Hardt, M., Levine, S. (eds.) Advances in Neural Information Processing Systems. vol.~36, pp. 25365--25389. Curran Associates, Inc. (2023), \url{https://proceedings.neurips.cc/paper_files/paper/2023/file/4ff83037e8d97b2171b2d3e96cb8e677-Paper-Conference.pdf}

\bibitem{chen2025personavectorsmonitoringcontrolling}
Chen, R., Arditi, A., Sleight, H., Evans, O., Lindsey, J.: Persona vectors: Monitoring and controlling character traits in language models (2025), \url{https://arxiv.org/abs/2507.21509}

\bibitem{cheng2024conditional}
Cheng, C.H., Ruess, H., Wu, C., Zhao, X.: Conditional fairness for generative ais. arXiv preprint arXiv:2404.16663  (2024)

\bibitem{chuang2023debiasing}
Chuang, C.Y., Jampani, V., Li, Y., Torralba, A., Jegelka, S.: Debiasing vision-language models via biased prompts. arXiv preprint arXiv:2302.00070  (2023)

\bibitem{couairon2022embedding}
Couairon, G., Douze, M., Cord, M., Schwenk, H.: Embedding arithmetic of multimodal queries for image retrieval. In: Proceedings of the IEEE/CVF Conference on Computer Vision and Pattern Recognition. pp. 4950--4958 (2022)

\bibitem{esser2024scaling}
Esser, P., Kulal, S., Blattmann, A., Entezari, R., M{\"u}ller, J., Saini, H., Levi, Y., Lorenz, D., Sauer, A., Boesel, F., et~al.: Scaling rectified flow transformers for high-resolution image synthesis. In: Forty-first international conference on machine learning (2024)

\bibitem{friedrich2025auditing}
Friedrich, F., Brack, M., Struppek, L., Hintersdorf, D., Schramowski, P., Luccioni, S., Kersting, K.: Auditing and instructing text-to-image generation models on fairness. AI and Ethics  \textbf{5}(3),  2103--2123 (2025)

\bibitem{gandikota2024concept}
Gandikota, R., Materzy{\'n}ska, J., Zhou, T., Torralba, A., Bau, D.: Concept sliders: Lora adaptors for precise control in diffusion models. In: European Conference on Computer Vision. pp. 172--188. Springer (2024)

\bibitem{he2016deep}
He, K., Zhang, X., Ren, S., Sun, J.: Deep residual learning for image recognition. In: Proceedings of the IEEE conference on computer vision and pattern recognition. pp. 770--778 (2016)

\bibitem{hessel-etal-2021-clipscore}
Hessel, J., Holtzman, A., Forbes, M., Le~Bras, R., Choi, Y.: {CLIPS}core: A reference-free evaluation metric for image captioning. In: Moens, M.F., Huang, X., Specia, L., Yih, S.W.t. (eds.) Proceedings of the 2021 Conference on Empirical Methods in Natural Language Processing. pp. 7514--7528. Association for Computational Linguistics, Online and Punta Cana, Dominican Republic (Nov 2021). \doi{10.18653/v1/2021.emnlp-main.595}, \url{https://aclanthology.org/2021.emnlp-main.595/}

\bibitem{hirotasaner}
Hirota, Y., Chen, M.H., Wang, C.Y., Nakashima, Y., Wang, Y.C.F., Hachiuma, R.: Saner: Annotation-free societal attribute neutralizer for debiasing clip. In: The Thirteenth International Conference on Learning Representations (2025)

\bibitem{hu2022lora}
Hu, E.J., Shen, Y., Wallis, P., Allen-Zhu, Z., Li, Y., Wang, S., Wang, L., Chen, W., et~al.: Lora: Low-rank adaptation of large language models. ICLR  \textbf{1}(2), ~3 (2022)

\bibitem{kang2025fairgen}
Kang, M., Kumar, V.B., Roy, S., Kumar, A., Khosla, S., Narayanaswamy, B.M., Gangadharaiah, R.: Fairgen: Controlling sensitive attributes for fair generations in diffusion models via adaptive latent guidance. arXiv preprint arXiv:2503.01872  (2025)

\bibitem{karkkainenfairface}
Karkkainen, K., Joo, J.: Fairface: Face attribute dataset for balanced race, gender, and age for bias measurement and mitigation. In: Proceedings of the IEEE/CVF Winter Conference on Applications of Computer Vision. pp. 1548--1558 (2021)

\bibitem{kim2025rethinking}
Kim, E., Kim, S., Park, M., Entezari, R., Yoon, S.: Rethinking training for de-biasing text-to-image generation: Unlocking the potential of stable diffusion. In: Proceedings of the Computer Vision and Pattern Recognition Conference. pp. 13361--13370 (2025)

\bibitem{kim2023stereotyping}
Kim, E., Kim, S., Shin, C., Yoon, S.: De-stereotyping text-to-image models through prompt tuning  (2023)

\bibitem{kim2021vilt}
Kim, W., Son, B., Kim, I.: Vilt: Vision-and-language transformer without convolution or region supervision (2021)

\bibitem{li2024self}
Li, H., Shen, C., Torr, P., Tresp, V., Gu, J.: Self-discovering interpretable diffusion latent directions for responsible text-to-image generation. In: Proceedings of the IEEE/CVF Conference on Computer Vision and Pattern Recognition. pp. 12006--12016 (2024)

\bibitem{blip_vqa}
Li, J., Li, D., Xiong, C., Hoi, S.: Blip: Bootstrapping language-image pre-training for unified vision-language understanding and generation (2022). \doi{10.48550/ARXIV.2201.12086}, \url{https://arxiv.org/abs/2201.12086}

\bibitem{luccioni2023stable}
Luccioni, S., Akiki, C., Mitchell, M., Jernite, Y.: Stable bias: Evaluating societal representations in diffusion models. Advances in Neural Information Processing Systems  \textbf{36},  56338--56351 (2023)

\bibitem{luo2024versusdebias}
Luo, H., Deng, Z., Huang, H., Liu, X., Chen, R., Liu, Z.: Versusdebias: Universal zero-shot debiasing for text-to-image models via slm-based prompt engineering and generative adversary. arXiv preprint arXiv:2407.19524  (2024)

\bibitem{mikolov2013efficient}
Mikolov, T., Chen, K., Corrado, G., Dean, J.: Efficient estimation of word representations in vector space. arXiv preprint arXiv:1301.3781  (2013)

\bibitem{mirza2014conditional}
Mirza, M., Osindero, S.: Conditional generative adversarial nets. arXiv preprint arXiv:1411.1784  (2014)

\bibitem{oppenlaender2022creativity}
Oppenlaender, J.: The creativity of text-to-image generation. In: Proceedings of the 25th international academic mindtrek conference. pp. 192--202 (2022)

\bibitem{parihar2024balancing}
Parihar, R., Bhat, A., Basu, A., Mallick, S., Kundu, J.N., Babu, R.V.: Balancing act: distribution-guided debiasing in diffusion models. In: Proceedings of the IEEE/CVF conference on computer vision and pattern recognition. pp. 6668--6678 (2024)

\bibitem{t2ibiassurvey}
Prerak, S.: Addressing bias in text-to-image generation: A review of mitigation methods. In: 2024 Third International Conference on Smart Technologies and Systems for Next Generation Computing (ICSTSN). pp.~1--6 (2024). \doi{10.1109/ICSTSN61422.2024.10671230}

\bibitem{radford2021learning}
Radford, A., Kim, J.W., Hallacy, C., Ramesh, A., Goh, G., Agarwal, S., Sastry, G., Askell, A., Mishkin, P., Clark, J., et~al.: Learning transferable visual models from natural language supervision. In: International conference on machine learning. pp. 8748--8763. PmLR (2021)

\bibitem{radford2015unsupervised}
Radford, A., Metz, L., Chintala, S.: Unsupervised representation learning with deep convolutional generative adversarial networks. arXiv preprint arXiv:1511.06434  (2015)

\bibitem{2020t5}
Raffel, C., Shazeer, N., Roberts, A., Lee, K., Narang, S., Matena, M., Zhou, Y., Li, W., Liu, P.J.: Exploring the limits of transfer learning with a unified text-to-text transformer. Journal of Machine Learning Research  \textbf{21}(140),  1--67 (2020), \url{http://jmlr.org/papers/v21/20-074.html}

\bibitem{rombach2022high}
Rombach, R., Blattmann, A., Lorenz, D., Esser, P., Ommer, B.: High-resolution image synthesis with latent diffusion models. In: Proceedings of the IEEE/CVF conference on computer vision and pattern recognition. pp. 10684--10695 (2022)

\bibitem{schroff2015facenet}
Schroff, F., Kalenichenko, D., Philbin, J.: Facenet: A unified embedding for face recognition and clustering. In: Proceedings of the IEEE conference on computer vision and pattern recognition. pp. 815--823 (2015)

\bibitem{serengil2024lightface}
Serengil, S., Ozpinar, A.: A benchmark of facial recognition pipelines and co-usability performances of modules. Journal of Information Technologies  \textbf{17}(2),  95--107 (2024). \doi{10.17671/gazibtd.1399077}, \url{https://dergipark.org.tr/en/pub/gazibtd/issue/84331/1399077}

\bibitem{seshadri-etal-2024-bias}
Seshadri, P., Singh, S., Elazar, Y.: The bias amplification paradox in text-to-image generation. In: Duh, K., Gomez, H., Bethard, S. (eds.) Proceedings of the 2024 Conference of the North American Chapter of the Association for Computational Linguistics: Human Language Technologies (Volume 1: Long Papers). pp. 6367--6384. Association for Computational Linguistics, Mexico City, Mexico (Jun 2024). \doi{10.18653/v1/2024.naacl-long.353}, \url{https://aclanthology.org/2024.naacl-long.353/}

\bibitem{shen2023finetuning}
Shen, X., Du, C., Pang, T., Lin, M., Wong, Y., Kankanhalli, M.: Finetuning text-to-image diffusion models for fairness. arXiv preprint arXiv:2311.07604  (2023)

\bibitem{simonyan2014very}
Simonyan, K., Zisserman, A.: Very deep convolutional networks for large-scale image recognition. arXiv preprint arXiv:1409.1556  (2014)

\bibitem{wan2025maleceofemaleassistant}
Wan, Y., Chang, K.W.: The male ceo and the female assistant: Evaluation and mitigation of gender biases in text-to-image generation of dual subjects (2025), \url{https://arxiv.org/abs/2402.11089}

\bibitem{wan2024surveybiastexttoimagegeneration}
Wan, Y., Subramonian, A., Ovalle, A., Lin, Z., Suvarna, A., Chance, C., Bansal, H., Pattichis, R., Chang, K.W.: Survey of bias in text-to-image generation: Definition, evaluation, and mitigation (2024), \url{https://arxiv.org/abs/2404.01030}

\bibitem{wang2022gitgenerativeimagetotexttransformer}
Wang, J., Yang, Z., Hu, X., Li, L., Lin, K., Gan, Z., Liu, Z., Liu, C., Wang, L.: Git: A generative image-to-text transformer for vision and language (2022), \url{https://arxiv.org/abs/2205.14100}

\bibitem{wang2023concept}
Wang, Z., Gui, L., Negrea, J., Veitch, V.: Concept algebra for (score-based) text-controlled generative models. Advances in Neural Information Processing Systems  \textbf{36},  35331--35349 (2023)

\bibitem{Zhang_2023_ICCV}
Zhang, C., Chen, X., Chai, S., Wu, C.H., Lagun, D., Beeler, T., De~la Torre, F.: Iti-gen: Inclusive text-to-image generation. In: Proceedings of the IEEE/CVF International Conference on Computer Vision (ICCV). pp. 3969--3980 (October 2023)

\bibitem{zhang2017stackgan}
Zhang, H., Xu, T., Li, H., Zhang, S., Wang, X., Huang, X., Metaxas, D.N.: Stackgan: Text to photo-realistic image synthesis with stacked generative adversarial networks. In: Proceedings of the IEEE international conference on computer vision. pp. 5907--5915 (2017)

\end{thebibliography}
\clearpage
\appendix
\section*{Appendix}
\begin{table*}[h]
\centering
\caption{Performance comparison of DeepFace, FairFace, and CLIP on race and ethnicity classification on the curated dataset.}
\label{tab:face_metrics}
\begin{tabular}{lcccc|cccc}
\hline
         & \multicolumn{4}{c|}{Gender}                & \multicolumn{4}{c}{Race}                                                                                            \\ \hline
         & Accuracy & Precision & Recall & F1-Score & \multicolumn{1}{l}{Accuracy} & \multicolumn{1}{l}{Precision} & \multicolumn{1}{l}{Recall} & \multicolumn{1}{l}{F1-Score} \\ \hline
DeepFace & 95.71    & 93.14     & 98.78  & 95.88    & 88.36                        & 89.93                         & 88.38                      & 88.01                        \\
FairFace & 100      & 100       & 100    & 100      & \textbf{92.8}                & \textbf{94.15}                & \textbf{92.8}              & \textbf{92.78}               \\
CLIP     & 100      & 100       & 100    & 100      & 84.83                        & 85.51                         & 84.83                      & 84.37                        \\ \hline
\end{tabular}%
\end{table*}
\section{Performance of the Attribute Predictors}
\label{sec:fairface_performance}
\begin{figure}[h]
    \centering
    \includegraphics[width=\linewidth]{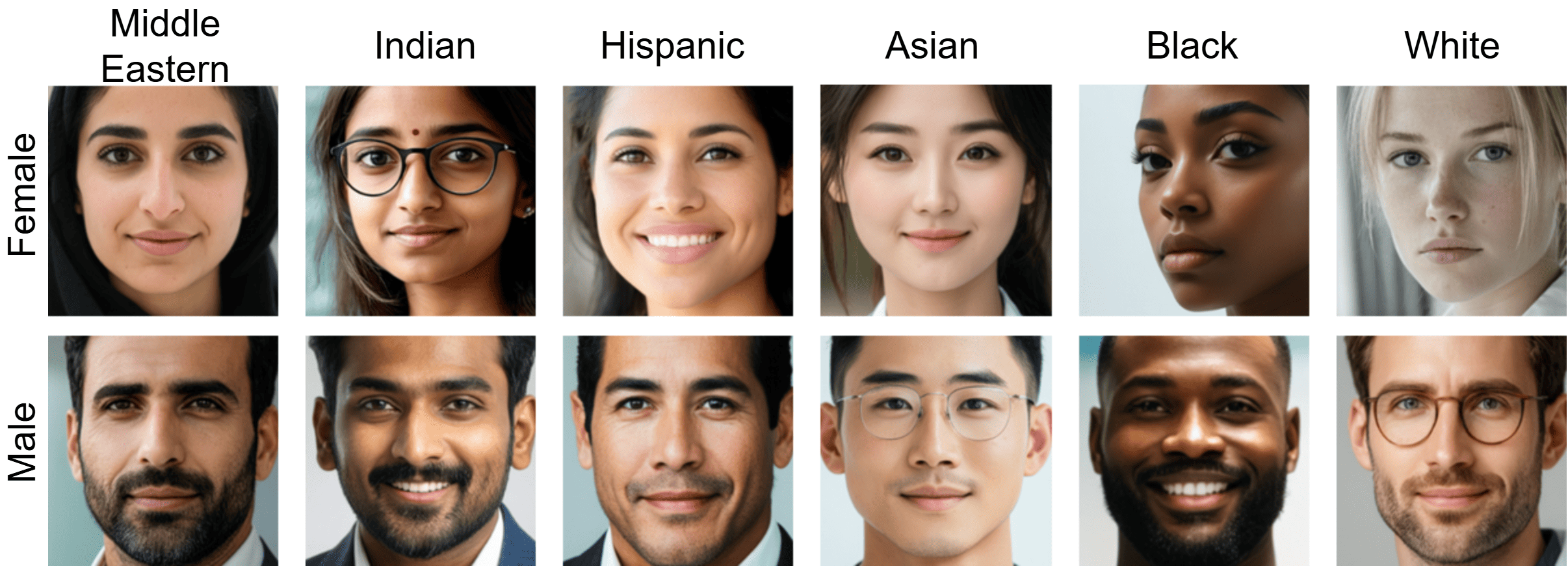}
    \caption{Samples from the curated dataset across all intersectional categories.}
    \label{fig:face_dataset}
\end{figure}

The gender and race attribute predictor is the crucial component of our analysis, Our design choice to select a ResNet \cite{he2016deep} based model trained on FairFace \cite{karkkainenfairface} dataset is based on the following experiment:

\paragraph{Dataset Curation}
The FairFace dataset, which was used to train the ResNet model has 2 gender categories \{male, female\} and 7 race categories \{white, black, south east asian, east asian, indian, latino/hispanic, middle eastern\}, for this analysis we only considered 6 categories combining south east asian and east asian.
We queried the FLUX T2I model to generate 100 portraits of people under all 12 intersectional categories (refer Figure \ref{fig:face_dataset}).
We then detected a bounding box over the faces in the portraits and then cropped them using the FaceNet algorithm \cite{schroff2015facenet} implemented on pytorch \footnote{\url{https://github.com/timesler/facenet-pytorch}}.

\paragraph{Model Choice}
We considered the following three models for comparison,
\begin{itemize}
    \item DeepFace Models\cite{serengil2024lightface} : are a collection of VGG architure\cite{simonyan2014very} based models trained on several datasets of face portraits, predicting the race and gender of a person in an image.
    \item FairFace Model \cite{karkkainenfairface} : is a ResNet based model that produces unified predictions of race and gender of the person in the given image.
    \item Zero-Shot CLIP Classifier \cite{radford2021learning} : is a Contrastively Learned Vision-Language transformer model, which can classify any given image into any arbitrary class based on the similarity scores between the image and text domain.
    We asked this model to classify the given images into the aforementioned gender and race categories.
\end{itemize}

To choose the best pre-trained architecture for our use case, we benchmarked all the models on the curated dataset.
We tracked the most relevant classification metrics like Accuracy, Precision, Recall and the F1-Score (Refer Table \ref{tab:face_metrics}).
We were able to observe that the FairFace and the CLIP based model were able to perfectly classify the genders in the images, while clip struggles to rightly classify the race, FairFace model continued to exhibit strong performance on the race classification task.

\begin{figure}
    \centering
    \includegraphics[width=0.7\linewidth]{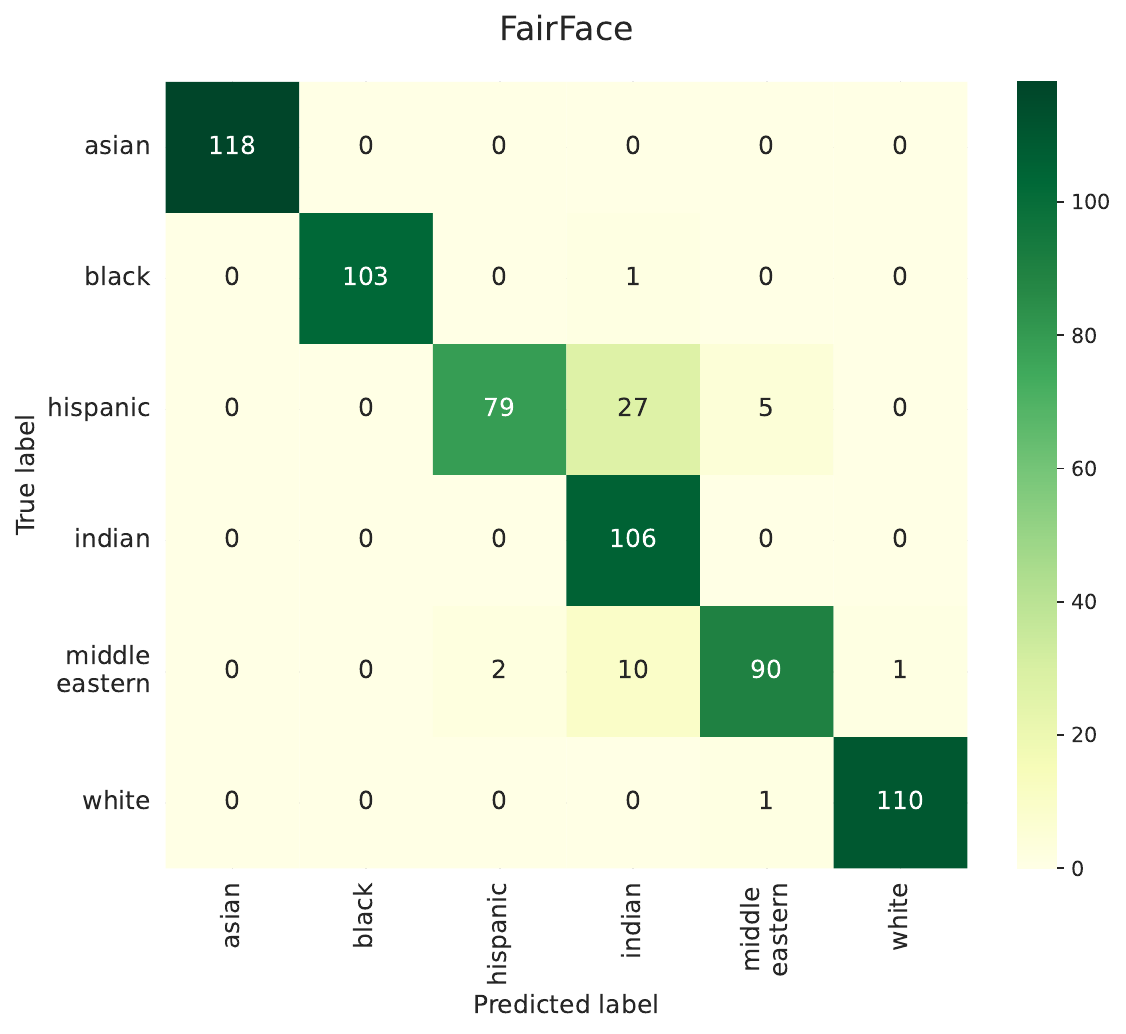}
    \caption{Confusion matrix of the predictions from the FairFace model on the race attributes}
    \label{fig:fairface_heatmap}
\end{figure}

Upon deeper analysis of the results from the FairFace model, we were able to observe that the model frequently misclassify Hispanic and Middle Eastern race to Indian (Refer  Figure \ref{fig:fairface_heatmap}).
Therefore, we choose to ignore these two race categories and only considered the following 4 race attributes : \{white, black, asian, indian\}.

\section{Implementation Details for Concept Coherence Score (CCS)}
\label{app:ccs_details}

Our Concept Coherence Score (CCS) is designed to be a robust and independent measure of a generated image's faithfulness to its base prompt. Here we detail the models, prompts, and method used to calculate CCS.

\paragraph{VQA Ensemble.} To reduce model-specific biases, we employ an ensemble of three pre-trained VQA models instead of a single one. The models used are BLIP-VQA-base~\cite{blip_vqa}, GIT-LARGE-TEXTVQA ~\cite{wang2022gitgenerativeimagetotexttransformer} and VILT-B-32~\cite{kim2021vilt}. 
Importantly, none of the VQA models in our ensemble share the same text encoder architecture (CLIP ViT-L/14) as our primary T2I models, ensuring independence between the generation and evaluation stages.

\paragraph{Prompting Strategy.} For each generated image and its corresponding base concept (e.g., "Nurse"), we prompt each VQA model in the ensemble with the following precise template: \texttt{Answer only with yes or no: Is the person in this image a \{concept\}?}
This strict template is designed to constrain the VQA models' output to a simple affirmative or negative response.

\begin{figure}[h]
    \centering
    \includegraphics[width=0.9\linewidth]{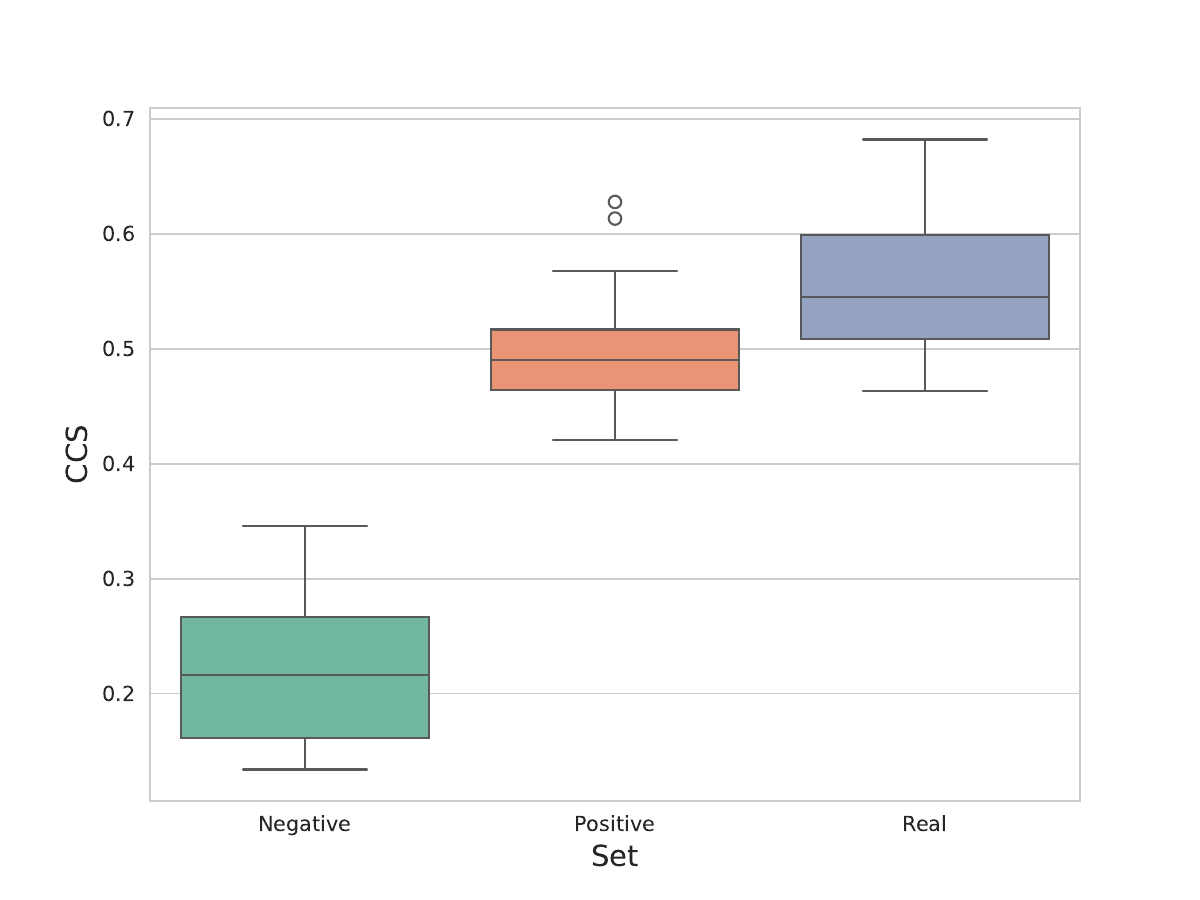}
    \caption{Box plots showing the CCS distributions for real-world images ('Real'), high-quality generated images ('Positives'), and unrelated images ('Negative') for the concept 'Carpenter'. The clear separation confirms that CCS effectively measures concept coherence.}
    \label{fig:ccs_validation}
\end{figure}

\paragraph{Score Calculation.} To move beyond binary predictions and create a continuous-valued score, we analyze the logits of the first output token from each VQA model. The CCS for a single image is the mean softmax probability assigned to the token \texttt{yes} across the $N$ models in the ensemble. Formally, it is calculated as:
\begin{equation}
\text{CCS} = \frac{1}{N} \sum_{i=1}^{N} P_{\theta_i}(\texttt{yes} \mid \text{image}, \text{concept})
\label{eq:ccs}
\end{equation}
where $P_{\theta_i}$ is the softmax probability from the $i$-th VQA model with parameters $\theta_i$. The score ranges from 0 to 1, where a higher value represents a stronger aggrement from the VQA ensemble that the image coherently represents the base concept. 
The final CCS reported in our results tables is the average score across all images generated for a given experimental condition.

\paragraph{Validation}
To validate our proposed CCS metric, we tested its ability to distinguish between different levels of concept coherence for the profession "Carpenter." We created three distinct image sets: a Real Set of images scraped from the internet, a Positive Set of high-quality images generated by our FLUX model, and a Negative Set of unrelated, random images. 
Each sets contains 30 images.
The results, visualized in Figure \ref{fig:ccs_validation}, confirm our hypothesis. The CCS distributions for the Real and Positive sets are both high and tightly clustered (median scores $> 0.45$ ), indicating the VQA ensemble is confident in their coherence. 
In contrast, the Negative Set received lower Coherence Score ($<$ 0.25).
The clear separation between these distributions demonstrates that our CCS metric is a reliable and valid measure of concept coherence, effectively mirroring the performance also on real-world data.

\section{Further Results}
\label{app:Further_results}
\begin{figure*}[t]
    \centering
    \includegraphics[width=\linewidth]{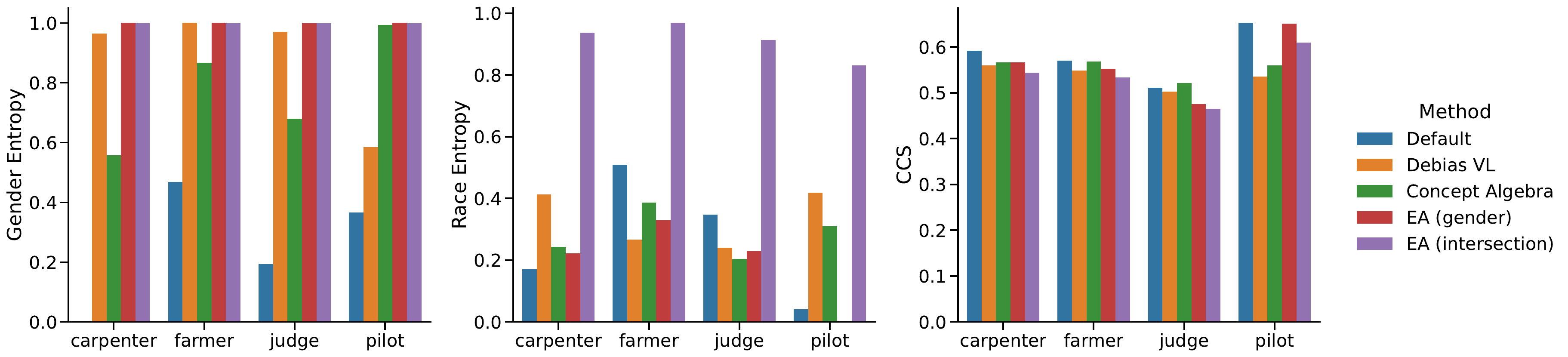}
    \caption{Comparison of gender entropy, race entropy, and concept consistency score (CCS) across professions under different debiasing methods.}
    \label{fig:more_quants}
\end{figure*}

\paragraph{Quantitative}
Apart from the concepts analysed in table \ref{tab:my-table}, we also sampled 4 more professions which could possess strong initial bias.
Similar to our generalization study, we queried the FLUX model to generate 100 images for each profession before and after applying the baselines and our bias mitigation methods.
Results for this experiments can be found in Figure \ref{fig:more_quants}

We were able to observe a similar pattern to our results in Section \ref{ssec:quant}, where the default model outputs continuously exhibited severe gender biases in several concepts like carpenter and judge, all the baseline methods as well as our $EA$ approach tends to significantly improve the diversity.
Furthermore our $EA_i$ approach also improves the diversity of the races in all the concepts, while also maintaining a CCS as close to the default.

\begin{figure}[h]
    \centering
    \begin{subfigure}[b]{0.48\textwidth}
        \centering
        \includegraphics[width=0.99\linewidth]{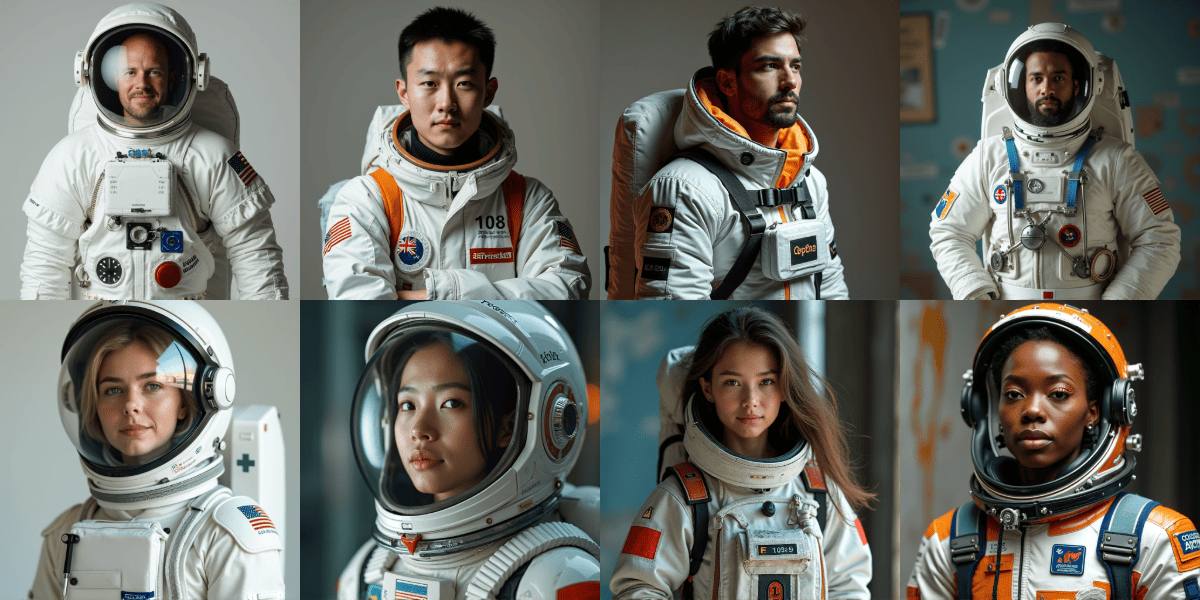}
        \caption{An \textbf{astronaut} posing for a photo with their suit on.}
        \label{fig:img1}
    \end{subfigure}
    \hfill
    \begin{subfigure}[b]{0.48\textwidth}
        \centering
        \includegraphics[width=0.99\linewidth]{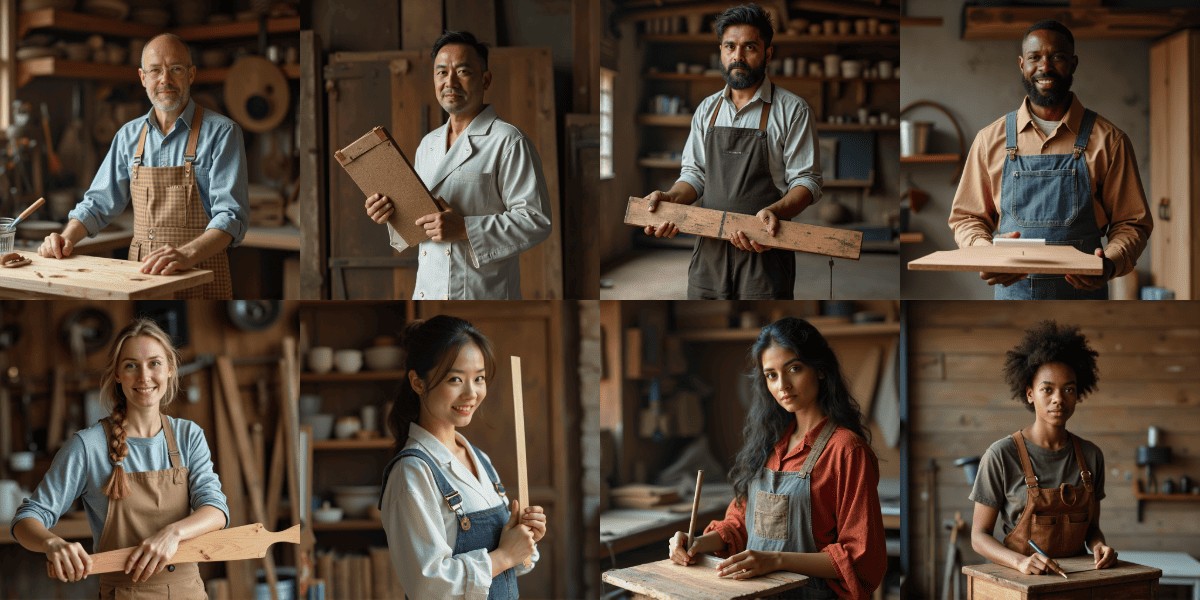}
        \caption{A photo of a \textbf{carpenter} in their workshop}
        \label{fig:img2}
    \end{subfigure}
    
    \vspace{0.5cm} 
    
    \begin{subfigure}[b]{0.48\textwidth}
        \centering
        \includegraphics[width=0.99\linewidth]{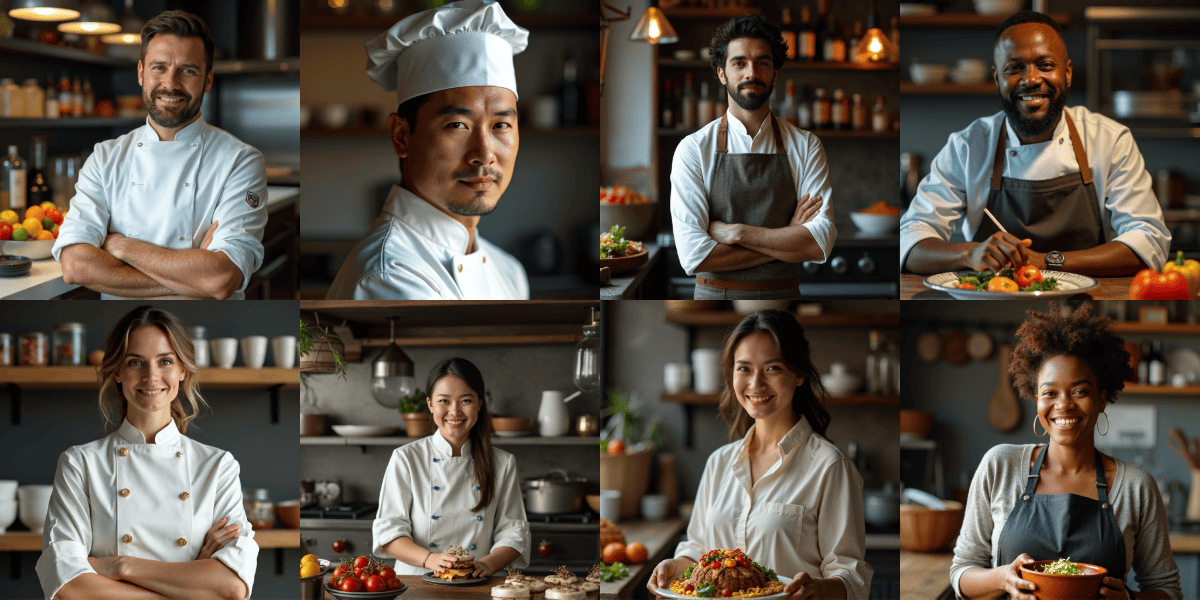}
        \caption{A photo of a \textbf{chef} from their kitchen.}
        \label{fig:img3}
    \end{subfigure}
    \hfill
    \begin{subfigure}[b]{0.48\textwidth}
        \centering
        \includegraphics[width=0.99\linewidth]{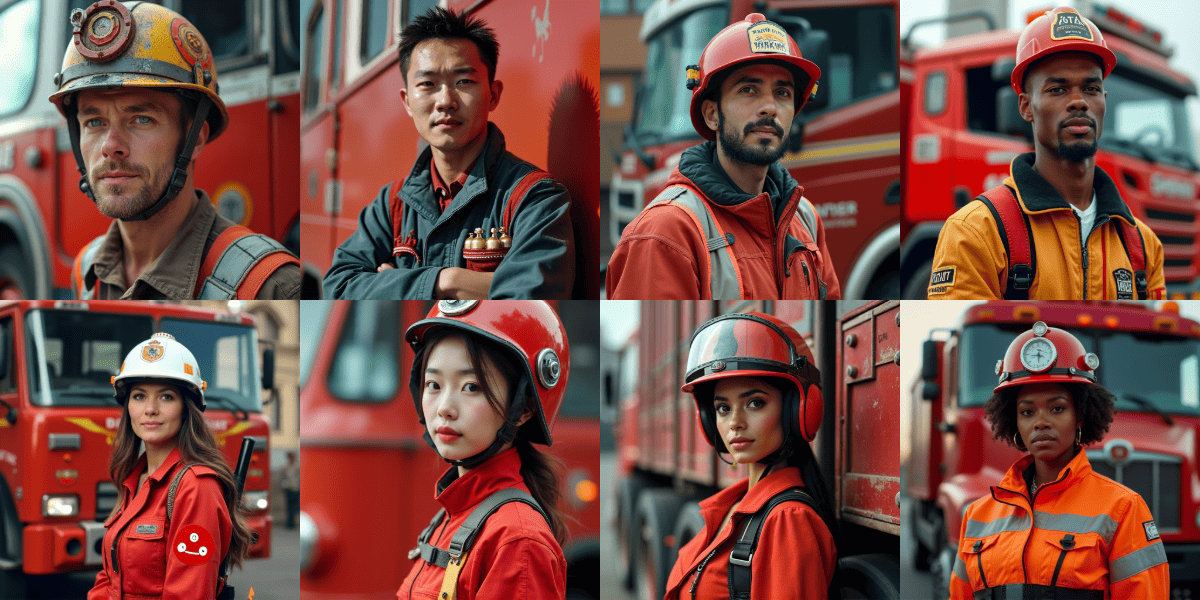}
        \caption{A photo of a \textbf{firefighter} posing besides a fire truck}
        \label{fig:img4}
    \end{subfigure}
    
    \caption{Images with diverse demographic attributes generated by the FLUX model for gender neutral prompts after applying our method.}
    \label{fig:qualitative_appendix}
\end{figure}

\paragraph{Qualitative}
To further validate the robustness and generalizability of our Embedding Arithmetic framework, we provide additional qualitative results in Figure \ref{fig:qualitative_appendix}.

These examples demonstrate two key properties. 
First, our intersectional method generalizes effectively to the unseen professions discussed in our main analysis, consistently generating diverse and coherent images for concepts with extreme baseline biases like "Firefighter" and "Carpenter."
Second, the results show that our method is not overfitted to the specific \texttt{a photo portrait of...} template used to create the attribute vectors; the steering remains effective even when applied to more varied and descriptive prompt styles.
This confirms that the derived attribute vectors are robust to changes in both the base concept and the linguistic structure of the prompt.

\begin{figure}[h]
    \centering
    \begin{subfigure}[b]{\textwidth}
        \centering
        \includegraphics[width=0.99\linewidth]{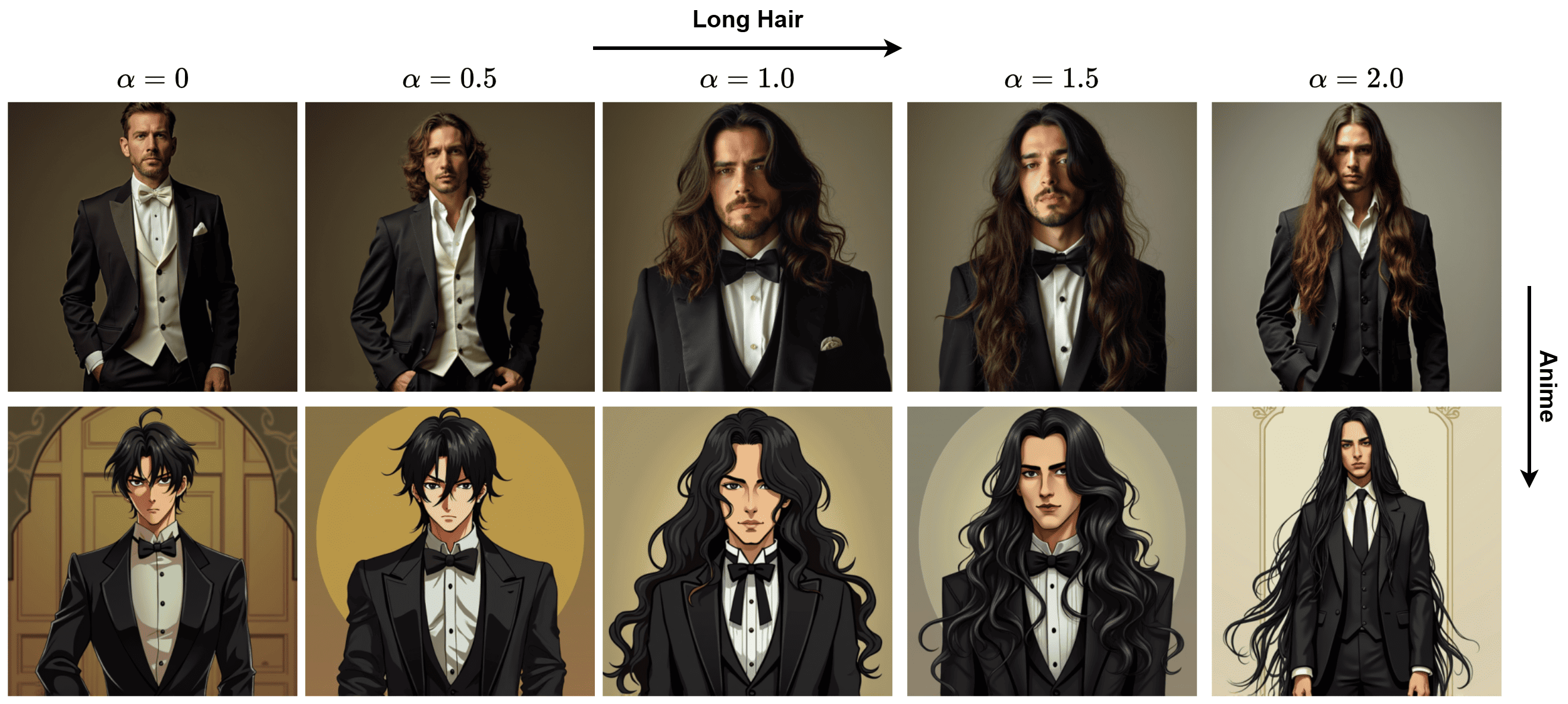}
        \caption{A photo of a man in a tuxedo}
        \label{fig:appendix_emotion}
    \end{subfigure}
    \hspace{0.5cm} 
    \begin{subfigure}[b]{\textwidth}
        \centering
        \includegraphics[width=0.99\linewidth]{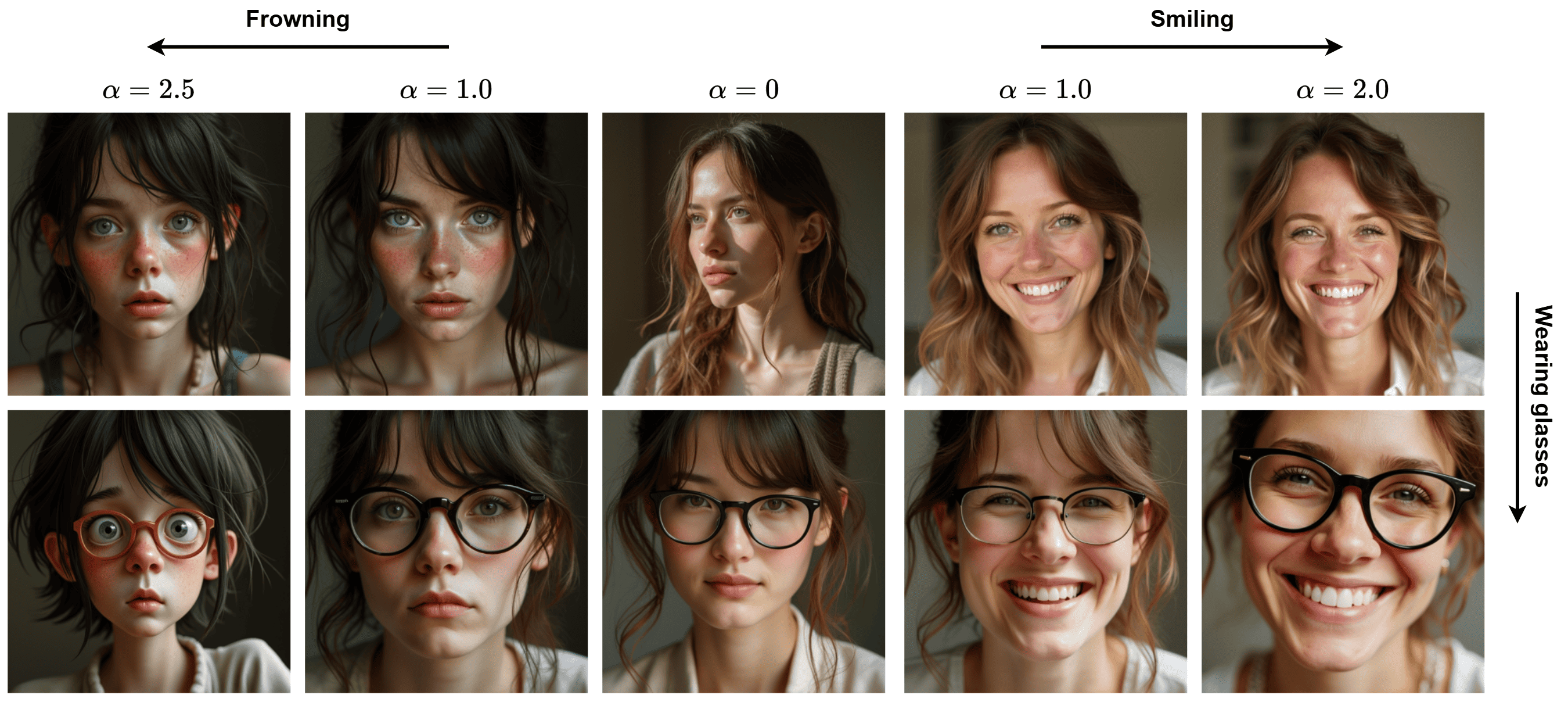}
        \caption{A close-up of a person.}
        \label{fig:appendix_hair}
    \end{subfigure}
    \caption{Images generated by the FLUX model for prompts mentioned in the sub-captions. The prompt embeddings are then moved toward the respective attributes using Embedding Arithmetic approach.}
\end{figure}

\section{Beyond Demographics}
\label{app:beyond_demographics}
To demonstrate that our framework is not limited to demographic attributes, we extended our approach to a diverse set of stylistic and visual concepts. 
We created new generalized attribute vectors for expressions (e.g., "smiling," "frowning"), hairstyles ("long hair"), accessories ("wearing glasses"), and artistic styles ("anime style"). 
These vectors were derived using the same methodology described in Section \ref{ssec:isolate}, by averaging the vector difference between a base prompt and a modified prompt (e.g., a photo portrait of a {profession}, smiling) across our set of source concepts. 
The qualitative results in Figures \ref{fig:appendix_emotion} and \ref{fig:appendix_hair} show that these non-demographic vectors can be successfully applied to new concepts, controllably adding the desired attributes while preserving the integrity of the base profession. 
This confirms that Embedding Arithmetic is a general-purpose framework for semantic manipulation in T2I models.

\pagebreak

\begin{figure}[h]
    \centering
        \centering
        \includegraphics[width=\linewidth]{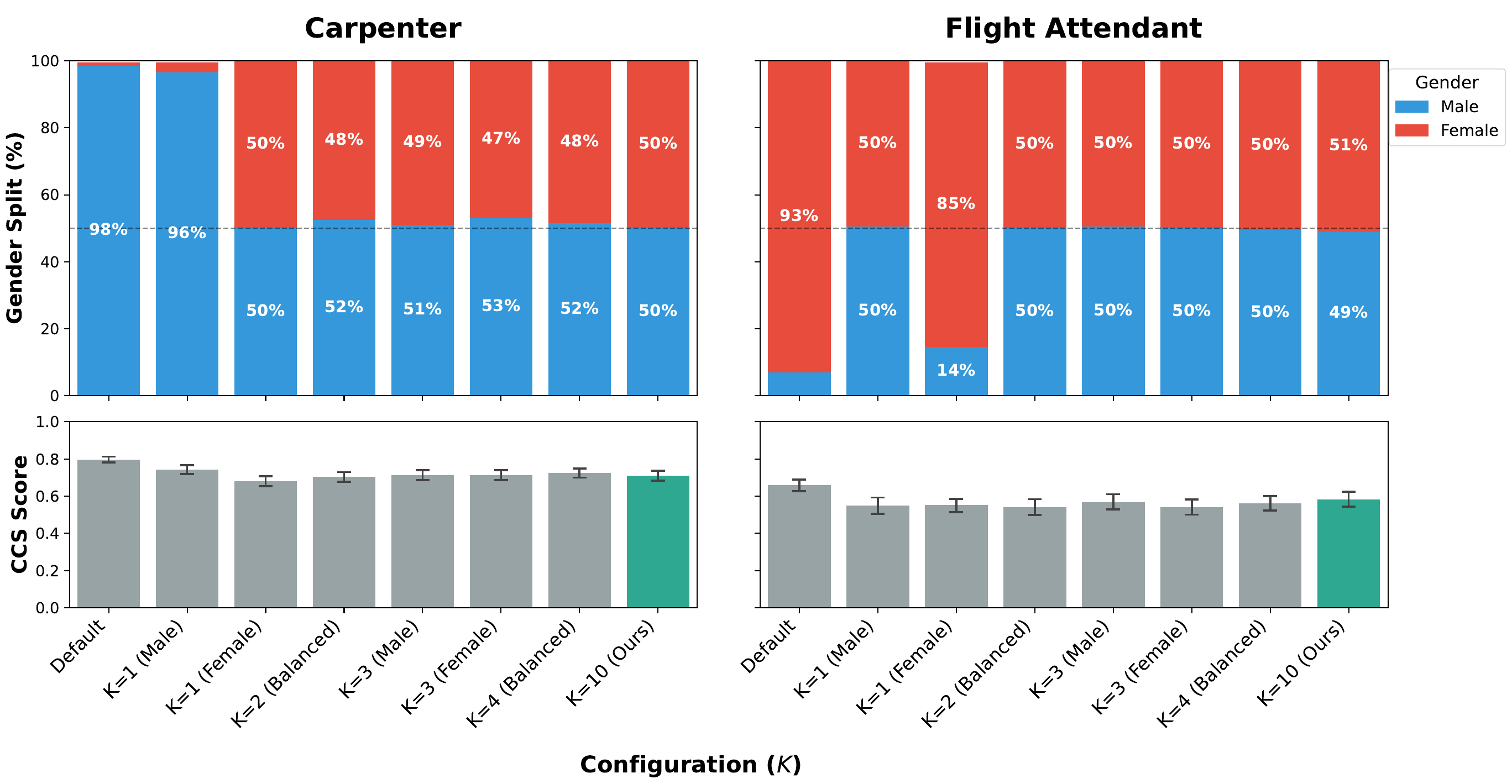}
        \caption{Gender Distributions (top row) and Concept Coherence Score (bottom row) for targets "Carpenter" (Left) and "Flight Attendant" (Column 2) genereted using various configurations of K. The results highlight the brittleness of using single-source vectors (K=1): utilizing a source profession that shares semantic priors with the target (e.g., using "Mechanic" as K=1 male source for "Carpenter") fails to mitigate bias (remaining 96\% Male). Conversly, while small balanced sets (K=2,4) achieve demographic parity, our full configuration (K = 10, green bar) consistently maintain the concept coherence compared to the default model.}
        \label{fig:k_ablation}
\end{figure}

\section{Ablation on the Base Concept Set Size (\textit{K})}
\label{sec:app_k_ablation}
Before characterizing the latent geometry of the text-conditioner in this work, we determined the optimal size and composition of the base concept set (\textit{K}) used to determine the attribute vectors in our Look-up table.
We hypothesized that vectors derived from single base concept (\textit{K} = 1) with strong initial bias towards a specific demography fail to capture the nuances of other demographics in the high dimensional Latent space.
To validate this we perform an ablation study (see Figure~\ref{fig:k_ablation}) on unseen targets that have strong initial bias like "Carpenter" (towards male) and "Flight Attendant" (towards female).
For each target concepts and each configuration of K parameter, we generated 200 images with different initial noise seed.

Our results confirm that using single-source concept (\textit{K} = 1) are brittle.
For instance, using a gender vector from male biased "Mechanic" (\textit{K} = 1) failed to debias the target "Carpenter" (still 96\% male), while using a gender vector from female biased "Childcare Worker" failed to debias the target "Flight Attendant" (still 85\%).

While increasing the diversity of base-concepts (K=4 with "Mechanic", "Childcare Worker", "Clergy" and "Interior Designer") rapidly stabilized the gender split, we found that a larget set of (K=10; refer Sec~\ref{ssec:setup}) yielded consistent Concept Coherence Score (CCS $\approx$ 0.71 for "Carpenters").
We posit that K=10 averages out residual profession-specific noise that could persist in smaller set.
Consequently, we adopt K = 10 for all subsequent analyses to ensure maximum vector generalizability.


\end{document}